\documentclass[10pt,twocolumn,letterpaper]{article}

\usepackage[pagenumbers]{iccv} 

\definecolor{iccvblue}{rgb}{0.21,0.49,0.74}
\usepackage{multirow}
\usepackage{pifont}
\usepackage{float}
\usepackage{algpseudocode}
\usepackage[ruled,vlined]{algorithm2e}
\usepackage[pagebackref,breaklinks,colorlinks,allcolors=iccvblue]{hyperref}
\usepackage[accsupp]{axessibility}  
\usepackage{array}
\usepackage{color}

\def\paperID{9496} 
\def\confName{ICCV}
\def\confYear{2025}

\title{CaliMatch: Adaptive Calibration for Improving Safe Semi-supervised Learning}

\author{
Jinsoo Bae\\
Korea University\\
Seoul, Republic of Korea
\and
Seoung Bum Kim\textsuperscript{*}\\
Korea University\\
Seoul, Republic of Korea
\and
Hyungrok Do\\
NYU Grossman School of Medicine\\
New York, NY, USA
}

\begin{document}
\maketitle
\def\thefootnote{*}\footnotetext{Seoung Bum Kim is the corresponding author: sbkim@korea.ac.kr.}
\begin{abstract}
    Semi-supervised learning (SSL) uses unlabeled data to improve the performance of machine learning models when labeled data is scarce. However, its real-world applications often face the label distribution mismatch problem, in which the unlabeled dataset includes instances whose ground-truth labels are absent from the labeled training dataset. Recent studies, referred to as safe SSL, have addressed this issue by using both classification and out-of-distribution (OOD) detection. However, the existing methods may suffer from overconfidence in deep neural networks, leading to increased SSL errors because of high confidence in incorrect pseudo-labels or OOD detection. To address this, we propose a novel method, CaliMatch, which calibrates both the classifier and the OOD detector to foster safe SSL. CaliMatch presents adaptive label smoothing and temperature scaling, which eliminates the need to manually tune the smoothing degree for effective calibration. We give a theoretical justification for why improving the calibration of both the classifier and the OOD detector is crucial in safe SSL. Extensive evaluations on CIFAR-10, CIFAR-100, SVHN, TinyImageNet, and ImageNet demonstrate that CaliMatch outperforms the existing methods in safe SSL tasks.
\end{abstract}
\section{Introduction}
Semi-supervised learning (SSL) is a useful approach to improving machine learning model performance, particularly when labeled data is limited, but plenty of unlabeled data are available \cite{kingma2014semi,saito2021openmatch,yoon2020vime,chen2023boosting,wei2023towards}. However, most existing SSL methods are based on the assumption that the labeled and unlabeled data have the same set of class labels. In reality, there could be unlabeled instances whose labels do not belong to any classes in the labeled dataset, called unseen-class data. Such unseen-class data can adversely affect the decision boundary of the classifier trained using labeled data and unlabeled seen-class data through SSL methods, thereby decreasing the effectiveness of SSL approaches. To address this issue, several deep learning-based SSL methods, known as \emph{safe semi-supervised learning}, have been proposed \cite{bae2022safe,chen2020semi,he2022safe,li2023iomatch,saito2021openmatch,yu2020multi}. Most safe SSL methods use out-of-distribution (OOD) detection techniques to identify whether the labels of unlabeled data belong to the set of labels in the labeled training dataset. That is, they categorize the unlabeled data into ``seen'' and ``unseen'' classes and use labeled data and unlabeled seen class data to improve the classifier or unlabeled unseen class data to improve the OOD detection performance.

Most safe SSL studies for image classification rely on deep convolutional neural networks (CNNs). However, recent studies have reported that deep CNNs show poor calibration performances despite their effectiveness measured in classification accuracies \cite{guo2017calibration,minderer2021revisiting,noh2023rankmixup,liu2022devil,tomani2021post}. Specifically, deep CNNs are overconfident in their decisions and struggle to accurately assess the likelihood of error based on their confidence levels. In this study, we investigate the overconfidence issue of deep CNNs in the context of safe SSL. The overconfidence in safe SSL can cause the classification model to learn from incorrect pseudo-labels with unreasonably high confidence in their decisions, thereby amplifying the model's error. Moreover, because of the overconfidence issue, OOD detection in safe SSL may erroneously label a considerable portion of unseen class data in the unlabeled dataset as a seen class, consequently undermining the effectiveness of safe SSL.

We propose a new safe SSL method, CaliMatch, to mitigate overconfidence-related issues in SSL by improving the calibration performance. CaliMatch effectively rejects unseen-class data and accurately identifies pseudo-labeled samples from the unlabeled dataset using a well-calibrated classifier and an OOD detector, thereby improving the efficacy of SSL. To achieve this, CaliMatch uses adaptive label smoothing and logit scaling to calibrate both the multiclass classifier and the OOD detector. Unlike conventional smoothing-based calibration methods, which require manual adjustment of the smoothing degree for effective calibration, CaliMatch dynamically adjusts the degree of label smoothing based on the accuracy distribution of the validation dataset, ensuring appropriate smoothing degrees relative to current confidence levels. Moreover, incorporating learnable scaling parameters for the multiclass classifier and OOD detector helps the models learn adaptively smoothed labels and mitigates overconfidence in both classification and OOD detection. This approach can alleviate the adverse training effects of incorrectly pseudo-labeled examples in safe SSL.

To the best of our knowledge, this is the first study to highlight the importance of improving the calibration performance of both multiclass classifiers and OOD detectors in safe SSL. The key contributions are as follows:
\begin{itemize}
    \item We investigate the importance of calibration in both classification and OOD detection in safe SSL, supported by extensive experiments and theoretical justification.
    \item We propose CaliMatch, a new safe SSL method that calibrates both the confidence of classifier and OOD scores to improve the quality of pseudo-label and effectively exclude instances from unseen classes within the unlabeled training dataset.
    \item CaliMatch outperforms existing safe SSL methods across five benchmark datasets, including a large-scale ImageNet dataset. Furthermore, our adaptive calibration approach is more effective than popular calibration methods in helping the safe SSL task.
\end{itemize}
\section{Related Works}
\textbf{Semi-supervised Learning.} Hybrid SSL methods integrate data augmentation and pseudo-labeling techniques to use large amounts of unlabeled data effectively \cite{berthelot2019mixmatch,2020remixmatch,sohn2020fixmatch,lee2013pseudo}. Notably, FixMatch has achieved state-of-the-art performance on several SSL benchmarks by using a confidence-based thresholding approach that considers pseudo-labels in training only when predictions meet a high confidence threshold. However, even with high confidence, pseudo-label quality can suffer because of overconfidence issues common in deep CNNs. To address this reliability concern, we propose calibrating the neural networks to improve the confidence-based thresholding process for accurately pseudo-labeled data, thus improving the quality of safe SSL.

\noindent \textbf{Safe Semi-supervised Learning.} 
Multi-task curriculum (MTC) trains an OOD detector alongside MixMatch using a joint optimization framework that alternates between updating neural network parameters and adjusting the OOD score \cite{yu2020multi}. SafeStudent uses a novel scoring function, known as energy discrepancy, to detect OOD instances \cite{he2022safe}. For better OOD detection, SafeStudent calibrates the probability distribution of detected unseen-class instances to a uniform distribution. OpenMatch, built on the framework of FixMatch, is an advanced, safe SSL method that incorporates an OOD rejection mechanism. IOMatch is based on OpenMatch and proposes to use multiclass and one-versus-rest (OvR) classifiers with a projection head to create an open-set classifier \cite{li2023iomatch}. Similarly, SCOMatch presents an open-set classifier and an OOD memory queue for selecting reliable OOD samples as new labeled data \cite{wang2024scomatch}. However, the overconfident neural network-based OOD scores may decrease the model's capability of correctly rejecting OOD instances with reliability during SSL. To address the negative issue in safe SSL, we propose calibrating an OOD detector and defining an OOD score by leveraging well-calibrated predictions of both the multiclass classifier and the OOD detector.

\noindent \textbf{Improving Calibration Performance of Deep Learning.} Calibration methods generally fall into two categories: (i) post-hoc techniques, such as temperature scaling \cite{guo2017calibration,liang2018enhancing}, which calibrate models after training, and (ii) real-time techniques that perform calibration during training. Our focus is on real-time calibration techniques because post-hoc calibration cannot correct for miscalibrated confidence that has already polluted the safe SSL process. Although smoothing-based calibration methods, such as mixup and label smoothing, can calibrate neural networks during training, achieving optimal calibration requires careful hyperparameter tuning to determine an appropriate degree of smoothing \cite{thulasidasan2019mixup,muller2019does}. To achieve efficient calibration for safe SSL, CaliMatch uses adaptive label smoothing based on the accuracy distribution over different confidence levels.


\noindent \textbf{Long-tailed Semi-supervised Learning.} Recent long-tailed SSL (LTSSL) methods are designed to address not only class imbalance in both labeled and unlabeled data, but also different imbalance ratios between them \cite{wei2023towards,sanchez2024flexible}. Specifically, these methods aim to correct biased pseudo-labels for unlabeled data, which are skewed toward majority classes. A recent LTSSL method, ADELLO \cite{sanchez2024flexible}, further incorporates the concept of calibration to mitigate overconfidence in majority classes and enhance the quality of pseudo-labels for minority classes. In contrast, CaliMatch addresses the presence of unlabeled samples from unknown classes, where calibration is critical for preventing overconfident pseudo-labeling of OOD data.
\begin{algorithm*}[!t]
\SetKwInput{KwIn}{Require}
\KwIn {$D_{\ell}$ and $D_{u}$: labeled and unlabeled datasets; $\phi_{\theta}$,  $f_{\theta}$, and $g_{\theta}$: encoder, multiclass classifier, and OOD detector; $T_M$ and $T_O$: learnable parameters; $\mathcal{T}_w$ and $\mathcal{T}_s$: weak and strong augmentations; $\tau_1$ and $\tau_2$: thresholds for consistency regularization and OOD rejection; $\lambda_{\text{O}},\lambda_{\text{OCal}}$, and $\lambda_{\text{S}}$: coefficients; $E_{max}$ and $I_{max}$: epoch and iteration; $E_{\text{warm-up}}$: warm-up epoch; $\eta$: learning rate}
 \For{$\text{Epoch} = 1$ to $E_{\text{max}}$}{
\For{$\text{Iteration} = 1$ to $I_{\text{max}}$}{
    Draw minibatches $B_{\ell}$ and $B_{u}$ from $D_{\ell}$ and $D_{u}$, respectively \\
    $\mathcal{L} = \mathcal{L}_{\text{CE}}(B_{\ell}) +\lambda_{\text{O}} \mathcal{L}_{\text{OOD}}(B_{\ell})  +\lambda_{\text{S}} \mathcal{L}_{\text{SC}}(B_{u}; \mathcal{T}_{w})$ \Comment{Equations \eqref{eq:cross-entropy}, \eqref{eq:binary cross-entropy}, \eqref{eq:soft-consistency}} \\
    \If{$\text{Epoch} \ge E_{\text{warm-up}}$}{
    $B_u^{t} = \{x_{i}^{u} \in B_{u} : (s_{i}^{u} > \tau_{1}) \wedge (c_{i}^{u} > \tau_{2})\}$ \Comment{Select unlabeled data with Equation \eqref{eq: CaliMatch condition}} \\
    $\mathcal{L} = \mathcal{L} + \mathcal{L}_{\text{MCal}}(B_\ell;\Gamma)+\lambda_{\text{OCal}}\mathcal{L}_{\text{OCal}}(B_\ell; \Delta) +\mathcal{L}_{\text{Fix}}(B_u^t;\mathcal{T}_w,\mathcal{T}_s) $ \Comment{Equations \eqref{eq:adaptive LS for classifier}, \eqref{eq:adaptive LS for OOD detector}, \eqref{eq:safe FixMatch}}
   }
   Update $[\theta,T_{M},T_{O}]$ at every iteration with stochastic gradient descent: $[\theta,T_{M},T_{O}] \gets [\theta,T_{M},T_{O}] - \eta \nabla \mathcal{L}$ \\
}
Update $\Gamma$ and $\Delta$ using validation dataset at every epoch\Comment{For adaptive label smoothing.}
}
\textbf{Output:} Trained networks for CaliMatch.
\caption{CaliMatch}
\label{alg1}
\end{algorithm*}

\section{Proposed Method}
\subsection{Safe Semi-supervised Learning}
In our problem setting, we are given a labeled dataset $D_{\ell} = \{(x_{i}^{\ell}, y_{i}^{\ell}) \in \mathcal{X} \times \mathcal{Y}:i=1,\cdots,n_{\ell}\}$, where $\mathcal{X}$ is the input feature space and $\mathcal{Y} = \{1,\cdots, K\}$ is the output variable space. Additionally, a set of unlabeled data $D_{u} = \{x_{i}^{u} \in \mathcal{X}:i=1,\cdots,n_{u}\}$ is given. We assume the presence of label distribution mismatch, where the true labels of the unlabeled instances in $D_{u}$ may not belong to $\mathcal{Y}$. Our primary goal is to build an accurate multiclass classification model using pseudo-labeling-based SSL techniques with both $D_{\ell}$ and $D_{u}$. However, to address the mismatch of the label distribution, we also require an OOD detector to distinguish between known classes (those in $\mathcal{Y}$) and unknown classes (those not in $\mathcal{Y}$). Therefore, following the safe SSL setting of \cite{saito2021openmatch}, we consider a neural network consisting of an encoder $\phi_{\theta}:\mathcal{X} \to \mathbb{R}^{d}$, which takes the input variable and generates a $d$-dimensional embedding vector that summarizes the features of each instance, a multiclass classifier $f_{\theta}: \mathbb{R}^{d} \to \mathbb{R}^{K}$, which takes an embedding vector as input and generates logit vectors for the classification task, and an OOD detector $g_{\theta}: \mathbb{R}^{d} \to \mathbb{R}^{K}$, implemented as a set of OvR binary classifiers, which generate logit vectors to distinguish unseen from seen classes.

Training a safe SSL model involves two key strategies: first, selecting a reliable subset of unlabeled data, which does not include instances of unseen classes from $D_u$ using the OOD detector, and second, applying consistency regularization techniques to the reliable subset. This reliable subset consists of unlabeled samples whose labels belong to the known classes, and their pseudo labels, derived from the model's predictions (i.e., $\text{argmax}_{k \in \mathcal{Y}} p_k(x_i^u)$), align with their true labels. To construct a reliable dataset, the existing safe SSL methods use OOD scores to discard unseen-class instances and use confidence scores in multiclass classification to correctly identify pseudo-labeled samples. However, because of the overconfidence issue of deep CNNs used in the existing safe SSL methods, the existing OOD and the confidence scores-based approach often fail to reject unlabeled instances successfully, thereby compromising the effectiveness of SSL techniques. Unexpectedly included unseen-class instances and incorrectly pseudo-labeled data can worsen a model's confirmation bias, resulting in poor classification performance. Therefore, addressing the overconfidence issue of deep CNN is crucial for the safe SSL task.

\subsection{CaliMatch: Safe Semi-supervised Learning with Improved Calibration}
To address the overconfident predictions of deep CNN and improve the efficacy of SSL, we propose an approach to improve the calibration performance of the multiclass classifier and the OOD detector by presenting adaptive label smoothing and temperature scaling techniques using labeled instances. Two scaling parameters $T_M$ and $T_O$ are initialized at 1.5 and optimized during training to allow the multiclass classifier and the OOD detector to learn adaptively smoothed labels efficiently. Our safe SSL method, CaliMatch, builds upon FixMatch, a highly effective consistency regularization SSL technique. CaliMatch incorporates an OOD detector to reject instances with unseen labels effectively and is trained to be well-calibrated, thereby enhancing the efficacy of consistency regularization.

Algorithm \ref{alg1} outlines CaliMatch. The encoder, multiclass classifier, and OOD detector are trained using a minibatch stochastic gradient descent. Given two sampled minibatches $B_{\ell}$ and $B_u$ from $D_{\ell}$ and $D_u$, respectively, in the warm-up phase, we update the model parameters to minimize three losses: cross-entropy loss for classification ($\mathcal{L}_{\text{CE}}(B_{\ell})$), binary cross-entropy loss for OOD ($\mathcal{L}_{\text{OOD}}(B_{\ell})$), and soft consistency regularization loss ($\mathcal{L}_{\text{SC}}(B_{u}; \mathcal{T}_{w})$) for the OOD detector. These three loss functions are given as follows:
{\small
\begin{align}
    -&\sum_{i=1}^{|B_{\ell}|} \sum_{k=1}^{K} y_{ik}^{\ell} \log p_{k}(x_{i}^{\ell}), \label{eq:cross-entropy}\\
    -&\sum_{i=1}^{|B_{\ell}|} \Big[ \sum_{k=1}^{K} \Big( y_{ik}^{\ell} \log q_{k}(x_{i}^{\ell}) \Big)+ \min_{ l \neq k} \log (1 - q_{l}(x_{i}^{\ell})) \Big], \label{eq:binary cross-entropy}\\
    &\sum_{i=1}^{|B_{u}|} \sum_{k=1}^{K} \Big( q_{k} \big( \mathcal{T}_{w}^{(1)}(x_{i}^{u}) \big) - q_{k}\big( \mathcal{T}_{w}^{(2)}(x_{i}^{u}) \big) \Big)^{2}, \label{eq:soft-consistency}
\end{align}}%
\noindent where $p(x_{i}^{\ell})$ is defined as $\text{softmax}((f_{\theta} \circ \phi_{\theta})(x_{i}^{\ell}))$ and $p_{k}(x_{i}^{\ell})$ denotes the $k$-th element of $p(x_{i}^{\ell})$, which represents the predicted probability of the instance $x_{i}^{\ell}$ being in class $k$. Note that $y_{ik}^{\ell}$ is equivalent to $\mathbb{I}(y_{i}^{\ell} = k)$. $q(x_{i}^{\ell})$ refers to $\text{sigmoid}((g_{\theta} \circ \phi_{\theta})(x_{i}^{\ell}))$, and $q_{k}(x_{i}^{\ell})$ denotes the $k$-th element of $q(x_{i}^{\ell})$, representing the predicted probability that $x_{i}^{\ell}$ is in class $k$. The distinction between the multiclass classifier and the OOD detector lies in how they interpret probabilities. Specifically, $p_{k}(x_{i}^{\ell})$ is the probability of being in class $k$, given that $x_{i}^{\ell}$ belongs to the in-distribution data. In contrast, $q_{k}(x_{i}^{\ell})$ indicates whether the instance belongs to class $k$ or not without conforming to the distribution of any known classes. $\mathcal{T}_{w}^{(1)}$ and $\mathcal{T}_{w}^{(2)}$ are transformations randomly selected from weak augmentation functions. The soft consistency regularization loss encourages the OOD detector to generate similar outputs for differently augmented input images, thereby improving OOD detection. 

After the warm-up period, we add the following two loss functions $\mathcal{L}_{\text{MCal}}(B_{\ell}; \Gamma)$ and $\mathcal{L}_{\text{OCal}}( B_{\ell}; \Delta)$ to improve calibration performance of the multiclass classifier and the OOD detector on the labeled data. These two loss functions are formulated as follows:
\begin{align}
    & -\sum_{i=1}^{|B_{\ell}|}\sum_{k=1}^{K} \Big( \gamma_{i}^{\ell} y_{ik}^{\ell} + \frac{1 - \gamma_{i}^{\ell}}{K-1} (1-y_{ik}^{\ell})\Big) \log p_{k}^{s}(x_{i}^{\ell}), \label{eq:adaptive LS for classifier} 
\end{align}
{\small
\begin{align}
    &-\sum_{i=1}^{|B_{\ell}|} \Big[ \sum_{k=1}^{K} \Big( \big(\delta_{i}^{\ell} y_{ik}^{\ell}+ (1-\delta_{i}^{\ell}) (1-y_{ik}^{\ell}) \big) \log q_{k}^{s}(x_{i}^{\ell}) \Big) \nonumber \\
    &+\underset{k \in \mathcal{Y}}{\min}{ \Big( \big((1-\delta_{i}^{\ell}) y_{ik}^{\ell} + \delta_{i}^{\ell} (1-y_{ik}^{\ell}) \big) \log (1 - q_{k}^{s}(x_{i}^{\ell})) \Big)} \Big], \label{eq:adaptive LS for OOD detector}
\end{align}}%
\noindent where $\Gamma$ and $\Delta$ are sets of reference confidence values for the multiclass classifier and the OOD detector, respectively. Specifically, $\gamma_{i}^{\ell}$ and $\delta_{i}^{\ell}$ are the reference confidence values assigned to the instance $x_{i}^{\ell}$ to determine the degrees of label smoothing based on the current model's confidence level. Unlike classic label smoothing, which applies a fixed smoothing degree to every sample, CaliMatch dynamically adjusts the smoothing levels for each instance to \(1-\gamma_{i}^{\ell}\) for the classifier and \(1-\delta_{i}^{\ell}\) for the OOD detector. Here, $p_{k}^{s}(x_{i}^{\ell})$ and $q_{k}^{s}(x_{i}^{\ell})$ denote the $k$-th elements of scaled-logit-based probabilities, $\text{softmax}((f_{\theta} \circ \phi_{\theta})(x_{i}^{\ell})/T_M)$ and $\text{softmax}((g_{\theta} \circ \phi_{\theta})(x_{i}^{\ell})/T_O)$, respectively. To assign reference value $\gamma_{i}^{\ell}$ for the multiclass classifier, we first divide the segment $[0, 1]$ into $M$ equally spaced bins and define a set of validation instances whose confidences belong to each segment: $B_{m}^{p} = \{x_{i} \in D_{\text{Val}}: \max_{k} p_{k}(x_{i}) \in (\frac{m-1}{M}, \frac{m}{M}] \}$ for $m=1,\cdots, M$. Subsequently, we calculate the accuracy of the multiclass classifier for each bin $B_{m}^{p}$ as $\gamma_{m} = |B_{m}^p|^{-1} \sum_{x_{i} \in B_{m}^p} \mathbb{I}(\text{argmax}_{k} p_{k}(x_{i}) = y_{i})$. Lastly, when we suppose the confidence value $\text{max}_{k} p_{k}(x_{i}^{\ell})$ for a labeled example $x_{i}^{\ell}$ falls into $B_{m_1}^p$, CaliMatch considers the accuracy value $\gamma_{m_1}$ as the well-calibrated reference confidence score $\gamma_{i}^{\ell}$ for the labeled data $x_{i}^{\ell}$. The same method calculates $\delta_{i}^{\ell}$ for the OOD detector using $B_m^q$. Through minimizing our calibration loss functions, the multiclass classifier and the OOD detector with two scaling parameters $T_M$ and $T_O$ learn the smoothed labels, thereby aligning their confidences $p_{y_i^\ell}^s(x_{i}^{\ell})$ and $q_{y_i^\ell}^s(x_{i}^{\ell})$ with the current model's accuracy $\gamma_{i}^{\ell}$ and $\delta_{i}^{\ell}$. Note that our smoothed labels take the reference values $\gamma_{i}^{\ell}$ and $\delta_{i}^{\ell}$ for the true class, and $(1-\gamma_{i}^{\ell})/(K-1)$ and $1-\delta_{i}^{\ell}$ for the remaining classes.

Next, we add FixMatch loss $\mathcal{L}_{\text{Fix}}(B_u^t;\mathcal{T}_w,\mathcal{T}_s)$ based on the set of reliable unlabeled instances $B_{u}^t$ for consistency regularization. The FixMatch loss is:
\begin{align}
    -\sum_{i=1}^{|B_{u}^t|}\sum_{k=1}^K \mathbb{I}\big(\text{argmax}_{l} p_{l}( \mathcal{T}_{w}(x_{i}^{u})) = k \big) \log p_{k}( \mathcal{T}_{s}(x_{i}^{u})), \label{eq:safe FixMatch}
\end{align}
where $\mathcal{T}_{w}$ and $\mathcal{T}_{s}$ are weak and strong augmentations applied to input variables, and they can be used to improve the classifier's consistency, thereby increasing the performance of the classifier, as used in FixMatch \citep{sohn2020fixmatch}. To identify the reliable set of unlabeled instances $B_{u}^t$ with classes in $\mathcal{Y}$ while ensuring the correctness of their pseudo-labels, we propose two selection criteria based on predictions of the model calibrated with the labeled dataset. The predictions of the model on the data $x_{i}^u \in B_{u}^t$ satisfy the following condition with two thresholds, $\tau_1$ and $\tau_2$:
\begin{equation}
s_i^u=\sum_{k \in \mathcal{Y}}p_k^s(x_i^u)q_k^s(x_i^u) > \tau_1 \quad \wedge \quad c_i^u=\max_{k \in \mathcal{Y}}p_k^s(x_i^u) > \tau_2. \label{eq: CaliMatch condition}
\end{equation}
Our proposed seen-class score $s_i^u$ consists of 
predicted probabilities $p_k^s(x_i^u)$ and $q_k^s(x_i^u)$, estimating the likelihood that the unlabeled data point $x_{i}^u$ belongs to one of the seen classes. This is based on the fact that the two well-calibrated probabilities, namely $p_k^s(x_i^u)$ and $q_k^s(x_i^u)$, account for the likelihood of $x_{i}^u$ being $k$-th seen class 
better than $p_k(x_i^u)$ and $q_k(x_i^u)$, which are likely to be overconfident. On the other hand, $u_i^u$, calculated as $1-s_i^u$, serves as a CaliMatch OOD score implying the likelihood that $x_{i}^u$ belongs to unseen classes not included in $\mathcal{Y}$. CaliMatch considers an unlabeled sample as reliable seen-class data if $s_i^u$ exceeds $\tau_1$. Furthermore, we propose using better-calibrated confidence $c_i^u$ to select unlabeled samples whose prediction confidence is greater than $\tau_2$ among those classified as unlabeled seen-class data, instead of using the overconfident score $\max_{k \in \mathcal{Y}}p_k(x_i^u)$ used in other safe SSL methods. Unless otherwise noted, $\tau_1$ and $\tau_2$ are fixed at 0.5 and 0.95, which are the same values from previous studies \cite{saito2021openmatch,sohn2020fixmatch}.
    
In Supplementary Section S-3, we present a theoretical justification demonstrating how better-calibrated models improve safe SSL performance. Specifically, we demonstrate that \emph{mitigating overconfidence in both classification and OOD detection can bring the gradient of $\mathcal{L}_{\text{Fix}}(B_u^t;\mathcal{T}_w,\mathcal{T}_s)$, closer to that of the ideal scenario (no incorrect pseudo-labels and no OOD samples) in safe SSL settings.}

\begin{table*}[]
\caption{Evaluation of multiclass classification using the averaged accuracy and standard deviation (in parentheses) on four image benchmark datasets under two different existence rates ($\kappa$\%) of unseen-label data. The best results are in \textbf{bold}, and the second-best results are \underline{underlined}.}
\label{table1}
\centering
\resizebox{0.8\linewidth}{!}{
\begin{tabular}{c|c|ccccccccc}
\hline
\multirow{2}{*}{Dataset}      & \multirow{2}{*}{$\kappa$} & \multicolumn{9}{c}{Method}                                                                                                                                                                                                                                                                                                                                                                                                                                                                                                                                                                                                                                                                                                                \\ \cline{3-11} 
                              &                           & \multicolumn{1}{c|}{Supervised}                                                              & \multicolumn{1}{c|}{MTC}                                                    & \multicolumn{1}{c|}{FixMatch}                                               & \multicolumn{1}{c|}{OpenMatch}                                              & \multicolumn{1}{c|}{SafeStudent}                                            & \multicolumn{1}{c|}{IOMatch}                                                & \multicolumn{1}{c|}{SCOMatch}                                                        & \multicolumn{1}{c|}{ADELLO}                                                 & \textbf{CaliMatch}                                              \\ \hline
\multirow{2}{*}{\raisebox{-3mm}{SVHN}}         & 30\%                      & \multicolumn{1}{c|}{\multirow{2}{*}{\begin{tabular}[c]{@{}c@{}}\raisebox{-3mm}{85.82}\\ (1.24)\end{tabular}}} & \multicolumn{1}{c|}{\begin{tabular}[c]{@{}c@{}}89.34\\ (2.09)\end{tabular}} & \multicolumn{1}{c|}{\begin{tabular}[c]{@{}c@{}}95.33\\ (0.38)\end{tabular}} & \multicolumn{1}{c|}{\begin{tabular}[c]{@{}c@{}}\underline{96.62}\\ \underline{(0.23)}\end{tabular}} & \multicolumn{1}{c|}{\begin{tabular}[c]{@{}c@{}}90.94\\ (0.62)\end{tabular}} & \multicolumn{1}{c|}{\begin{tabular}[c]{@{}c@{}}95.72\\ (0.43)\end{tabular}} & \multicolumn{1}{c|}{\begin{tabular}[c]{@{}c@{}}95.80\\ (0.48)\end{tabular}}          & \multicolumn{1}{c|}{\begin{tabular}[c]{@{}c@{}}95.70\\ (0.51)\end{tabular}} & \textbf{\begin{tabular}[c]{@{}c@{}}96.81\\ (0.23)\end{tabular}} \\ \cline{2-2} \cline{4-11} 
                              & 60\%                      & \multicolumn{1}{c|}{}                                                                        & \multicolumn{1}{c|}{\begin{tabular}[c]{@{}c@{}}86.77\\ (2.76)\end{tabular}} & \multicolumn{1}{c|}{\begin{tabular}[c]{@{}c@{}}94.20\\ (0.32)\end{tabular}} & \multicolumn{1}{c|}{\begin{tabular}[c]{@{}c@{}}\underline{95.67}\\ \underline{(0.14)}\end{tabular}} & \multicolumn{1}{c|}{\begin{tabular}[c]{@{}c@{}}90.34\\ (0.49)\end{tabular}} & \multicolumn{1}{c|}{\begin{tabular}[c]{@{}c@{}}94.52\\ (0.32)\end{tabular}} & \multicolumn{1}{c|}{\begin{tabular}[c]{@{}c@{}}94.59\\ (0.38)\end{tabular}}          & \multicolumn{1}{c|}{\begin{tabular}[c]{@{}c@{}}94.30\\ (0.43)\end{tabular}} & \textbf{\begin{tabular}[c]{@{}c@{}}96.56\\ (0.11)\end{tabular}} \\ \hline
\multirow{2}{*}{\raisebox{-3mm}{CIFAR-10}}     & 30\%                      & \multicolumn{1}{c|}{\multirow{2}{*}{\begin{tabular}[c]{@{}c@{}}\raisebox{-3mm}{76.96}\\ (0.39)\end{tabular}}} & \multicolumn{1}{c|}{\begin{tabular}[c]{@{}c@{}}82.22\\ (0.88)\end{tabular}} & \multicolumn{1}{c|}{\begin{tabular}[c]{@{}c@{}}88.12\\ (0.39)\end{tabular}} & \multicolumn{1}{c|}{\begin{tabular}[c]{@{}c@{}}88.04\\ (0.24)\end{tabular}} & \multicolumn{1}{c|}{\begin{tabular}[c]{@{}c@{}}81.88\\ (0.65)\end{tabular}} & \multicolumn{1}{c|}{\begin{tabular}[c]{@{}c@{}}89.12\\ (0.31)\end{tabular}} & \multicolumn{1}{c|}{\textbf{\begin{tabular}[c]{@{}c@{}}90.30\\ (0.35)\end{tabular}}} & \multicolumn{1}{c|}{\begin{tabular}[c]{@{}c@{}}88.81\\ (0.31)\end{tabular}} & \begin{tabular}[c]{@{}c@{}}\underline{90.25}\\ \underline{(0.27)}\end{tabular}          \\ \cline{2-2} \cline{4-11} 
                              & 60\%                      & \multicolumn{1}{c|}{}                                                                        & \multicolumn{1}{c|}{\begin{tabular}[c]{@{}c@{}}79.74\\ (0.42)\end{tabular}} & \multicolumn{1}{c|}{\begin{tabular}[c]{@{}c@{}}85.52\\ (0.54)\end{tabular}} & \multicolumn{1}{c|}{\begin{tabular}[c]{@{}c@{}}86.19\\ (0.74)\end{tabular}} & \multicolumn{1}{c|}{\begin{tabular}[c]{@{}c@{}}80.64\\ (0.33)\end{tabular}} & \multicolumn{1}{c|}{\begin{tabular}[c]{@{}c@{}}86.23\\ (0.92)\end{tabular}} & \multicolumn{1}{c|}{\begin{tabular}[c]{@{}c@{}}\underline{86.80}\\ \underline{(1.00)}\end{tabular}}          & \multicolumn{1}{c|}{\begin{tabular}[c]{@{}c@{}}86.73\\ (0.78)\end{tabular}} & \textbf{\begin{tabular}[c]{@{}c@{}}87.62\\ (0.36)\end{tabular}} \\ \hline
\multirow{2}{*}{\raisebox{-3mm}{CIFAR-100}}    & 30\%                      & \multicolumn{1}{c|}{\multirow{2}{*}{\begin{tabular}[c]{@{}c@{}}\raisebox{-3mm}{60.24}\\ (0.53)\end{tabular}}} & \multicolumn{1}{c|}{\begin{tabular}[c]{@{}c@{}}66.88\\ (0.68)\end{tabular}} & \multicolumn{1}{c|}{\begin{tabular}[c]{@{}c@{}}68.74\\ (0.48)\end{tabular}} & \multicolumn{1}{c|}{\begin{tabular}[c]{@{}c@{}}\underline{69.65}\\ \underline{(0.93)}\end{tabular}} & \multicolumn{1}{c|}{\begin{tabular}[c]{@{}c@{}}64.38\\ (1.00)\end{tabular}} & \multicolumn{1}{c|}{\begin{tabular}[c]{@{}c@{}}69.35\\ (0.78)\end{tabular}} & \multicolumn{1}{c|}{\begin{tabular}[c]{@{}c@{}}66.33\\ (0.64)\end{tabular}}          & \multicolumn{1}{c|}{\begin{tabular}[c]{@{}c@{}}69.00\\ (0.96)\end{tabular}} & \textbf{\begin{tabular}[c]{@{}c@{}}72.32\\ (0.34)\end{tabular}} \\ \cline{2-2} \cline{4-11} 
                              & 60\%                      & \multicolumn{1}{c|}{}                                                                        & \multicolumn{1}{c|}{\begin{tabular}[c]{@{}c@{}}63.02\\ (0.64)\end{tabular}} & \multicolumn{1}{c|}{\begin{tabular}[c]{@{}c@{}}64.66\\ (0.53)\end{tabular}} & \multicolumn{1}{c|}{\begin{tabular}[c]{@{}c@{}}\underline{65.65}\\ \underline{(1.02)}\end{tabular}} & \multicolumn{1}{c|}{\begin{tabular}[c]{@{}c@{}}62.72\\ (1.19)\end{tabular}} & \multicolumn{1}{c|}{\begin{tabular}[c]{@{}c@{}}65.59\\ (0.81)\end{tabular}} & \multicolumn{1}{c|}{\begin{tabular}[c]{@{}c@{}}63.78\\ (0.91)\end{tabular}}          & \multicolumn{1}{c|}{\begin{tabular}[c]{@{}c@{}}65.18\\ (0.28)\end{tabular}} & \textbf{\begin{tabular}[c]{@{}c@{}}68.56\\ (1.08)\end{tabular}} \\ \hline
\multirow{2}{*}{\raisebox{-3mm}{TinyImageNet}} & 30\%                      & \multicolumn{1}{c|}{\multirow{2}{*}{\begin{tabular}[c]{@{}c@{}}\raisebox{-3mm}{39.68}\\ (0.68)\end{tabular}}} & \multicolumn{1}{c|}{\begin{tabular}[c]{@{}c@{}}44.50\\ (0.97)\end{tabular}} & \multicolumn{1}{c|}{\begin{tabular}[c]{@{}c@{}}46.06\\ (0.67)\end{tabular}} & \multicolumn{1}{c|}{\begin{tabular}[c]{@{}c@{}}\underline{46.90}\\ \underline{(0.63)}\end{tabular}} & \multicolumn{1}{c|}{\begin{tabular}[c]{@{}c@{}}42.44\\ (0.38)\end{tabular}} & \multicolumn{1}{c|}{\begin{tabular}[c]{@{}c@{}}46.33\\ (0.89)\end{tabular}} & \multicolumn{1}{c|}{\begin{tabular}[c]{@{}c@{}}44.11\\ (0.89)\end{tabular}}          & \multicolumn{1}{c|}{\begin{tabular}[c]{@{}c@{}}44.72\\ (0.81)\end{tabular}}   & \textbf{\begin{tabular}[c]{@{}c@{}}47.46\\ (0.69)\end{tabular}} \\ \cline{2-2} \cline{4-11} 
                              & 60\%                      & \multicolumn{1}{c|}{}                                                                        & \multicolumn{1}{c|}{\begin{tabular}[c]{@{}c@{}}41.04\\ (0.89)\end{tabular}} & \multicolumn{1}{c|}{\begin{tabular}[c]{@{}c@{}}42.90\\ (0.54)\end{tabular}} & \multicolumn{1}{c|}{\begin{tabular}[c]{@{}c@{}}\underline{44.02}\\ \underline{(0.63)}\end{tabular}} & \multicolumn{1}{c|}{\begin{tabular}[c]{@{}c@{}}40.76\\ (0.83)\end{tabular}} & \multicolumn{1}{c|}{\begin{tabular}[c]{@{}c@{}}43.15\\ (0.95)\end{tabular}} & \multicolumn{1}{c|}{\begin{tabular}[c]{@{}c@{}}42.05\\ (0.83)\end{tabular}}          & \multicolumn{1}{c|}{\begin{tabular}[c]{@{}c@{}}42.80\\ (0.61)\end{tabular}} & \textbf{\begin{tabular}[c]{@{}c@{}}44.66\\ (0.37)\end{tabular}} \\ \hline
\end{tabular}
}
\end{table*}

\begin{figure*}[] 
\centering
\includegraphics[width=0.8\textwidth]{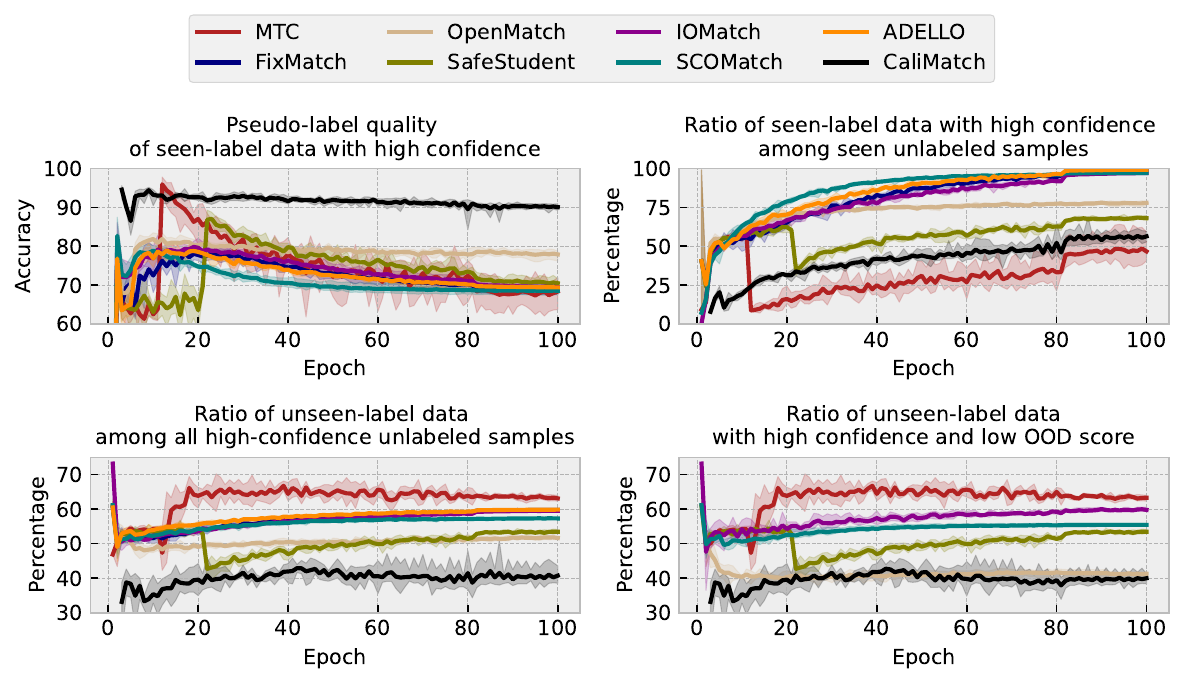}
\caption{Learning curves averaged over five runs on CIFAR-100 for CaliMatch and other SSL methods. The shaded region indicates standard deviations calculated from five runs.} \label{image1}
\end{figure*}

\section{Experiments}
\noindent \textbf{Setups.} We evaluated CaliMatch on a comprehensive list of benchmark datasets, including SVHN \cite{netzer2011reading}, CIFAR-10, CIFAR-100 \cite{krizhevsky2009learning}, TinyImageNet \cite{le2015tiny}, and ImageNet \cite{deng2009imagenet}, in the presence of label distribution mismatch between labeled and unlabeled datasets. We compared CaliMatch with conventional SSL, safe SSL methods, and a recent calibration-based LTSSL method. These include MTC \cite{yu2020multi}, FixMatch \cite{sohn2020fixmatch}, OpenMatch \cite{saito2021openmatch}, SafeStudent \cite{he2022safe}, IOMatch \cite{li2023iomatch}, SCOMatch \cite{wang2024scomatch}, and ADELLO \cite{sanchez2024flexible}. For fair evaluation, we followed the SSL evaluation protocol of \cite{oliver2018realistic}, which outlines the SSL training procedure, including the number of training iterations, CNN architecture, batch size, optimizer, learning rate, and scheduler. In experiments on SVHN, CIFAR-10, CIFAR-100, and TinyImageNet, we used a Wide ResNet 28-2 \cite{BMVC2016_87} as the CNN backbone, while ResNet 50 \cite{he2016deep} was used for experiments on ImageNet. All other implementation details, such as the settings of label distribution mismatch and hyperparameters in each dataset, are summarized in Supplementary Section S-1. Note that we also compared computational complexity of a supervised learning baseline and all SSL methods, and the results are presented in Supplementary Section S-2. To ensure robustness, we conducted experiments on SVHN, CIFAR-10, CIFAR-100, and TinyImageNet using the same evaluation protocols on five different random seeds, reporting average metrics and standard deviation. In the case of the ImageNet dataset, we limited the number of repeated experiments to three runs to alleviate extensive training costs. Lastly, we evaluated the calibration performance by calculating expected calibration error (ECE) with 15 bins \cite{naeini2015obtaining}. 

\subsection{Multiclass Classification}
Table \ref{table1} presents the classification performance (top-1 accuracy) across four image benchmark datasets, with unseen-label data proportion ($\kappa$) set to 30\% or 60\%. Across all eight combinations of datasets and proportions, five SSL methods improved performance compared to the supervised learning approach using only labeled data. Notably, the proposed CaliMatch outperformed most SSL competitors in both 30\% and 60\% proportions of unseen classes, demonstrating its efficacy and robustness in the presence of unlabeled OOD instances. Furthermore, a Friedman test conducted across the eight SSL scenarios yielded a statistically significant result ($p$-value $<$ 0.001), with CaliMatch ranked first based on the Friedman scores. We also evaluated the calibration performance of the methods, as presented in Supplementary Section S-2 (Table S-3). CaliMatch provides not only improved accuracy but also decent calibration performance. 

The proposed CaliMatch outperformed other methods by providing more accurate pseudo-labels for high-confidence instances (confidence score greater than 0.95), which are selected for consistency regularization. Additionally, it includes a lower proportion of unseen-label instances in the set of high-confidence and low-OOD instances (OOD score less than 0.5). To demonstrate this, we investigated the learning curves of the SSL methods. Figure \ref{image1} shows the learning curves for CIFAR-100 with $\kappa$ set to 60\%. Learning curves for other datasets are described in Supplementary Section S-2 (Figure S-1). First, as shown in the top-left figure, CaliMatch provides better pseudo-label accuracy for seen-class instances with confidence scores above the threshold. That is, the quality of the unlabeled instances selected by CaliMatch for consistency regularization is better than that of others, while the other SSL methods provide less accurate pseudo-labels. This advantage is because of CaliMatch's better calibration, whereas other methods suffer from overconfident predictions, as shown in Table S-3. This significantly improves the effectiveness of consistency regularization and leads CaliMatch to better classification performance.

The top-right figure depicts the ratio of high-confidence unlabeled samples with seen classes among all unlabeled samples with seen classes. While SCOMatch, ADELLO, and IOMatch showed high confidence in most of the training unlabeled data, they could not ensure the quality of pseudo-labels for high-confidence unlabeled data compared to CaliMatch. The bottom-left figure shows the ratio of unseen-label data among all high-confidence unlabeled samples. CaliMatch has the lowest percentage compared to all other comparison methods, indicating that it also improves OOD detection calibration. This leads to better detection of unseen classes in the high-confidence set, which is used for consistency regularization. Furthermore, CaliMatch's lower percentage in the bottom-right figure, indicating the proportion of unseen-label data among samples with low OOD scores and high confidence, highlights its superior capability to filter out unseen-label data in $D_u$.

\begin{figure*}[] 
\centering
\includegraphics[width=0.7\textwidth]{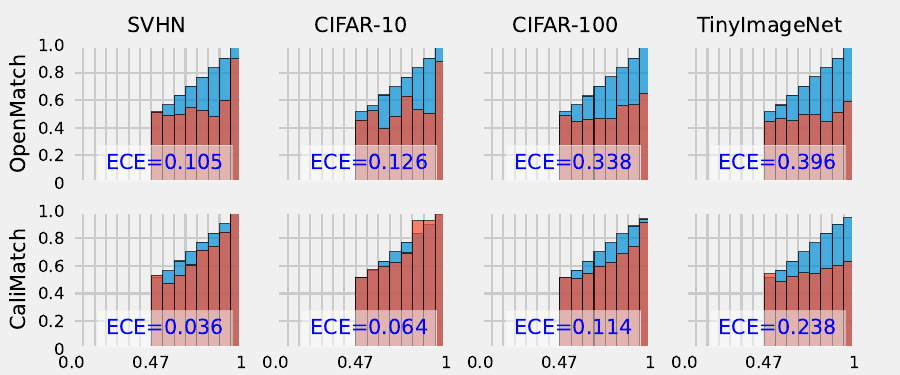}
\caption{Reliability diagrams of OpenMatch and CaliMatch for unseen-label data detection on four datasets. The blue and red bars represent the average confidence score and sample accuracy of the confidence bin, respectively.}
\label{image2}
\end{figure*}

\begin{figure*}[]
\centering
\includegraphics[width=0.7\linewidth]{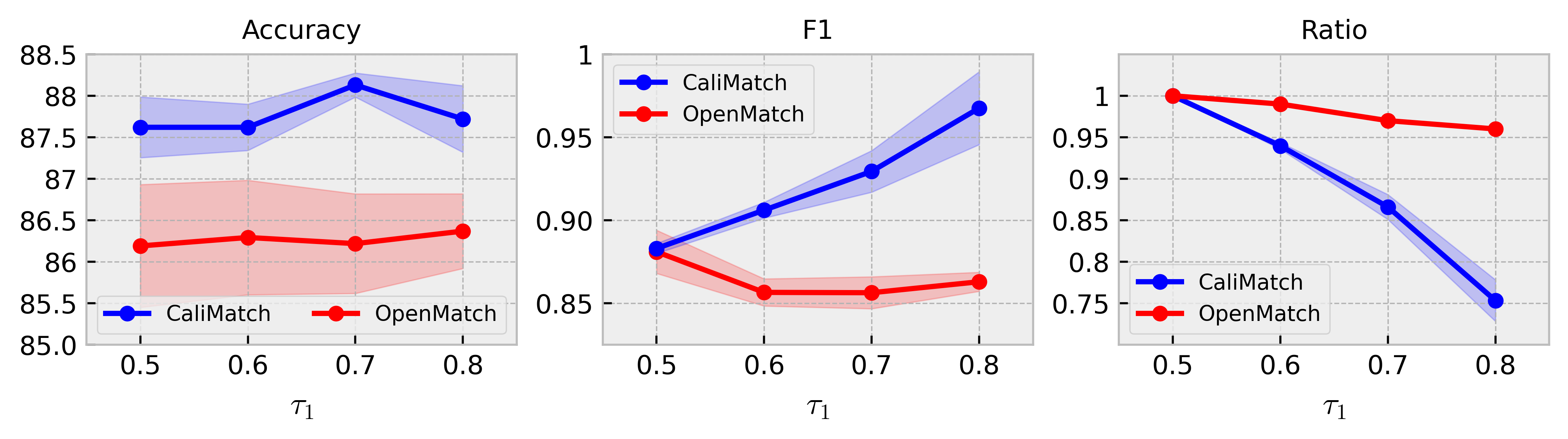}
\caption{Performance variations in CaliMatch and OpenMatch with threshold $\tau_{1}$ on CIFAR-10. We marked the averaged metrics and standard deviation as straight lines and shaded areas over five runs.} \label{image3}
\end{figure*}

\subsection{Out-of-distribution Detection}
We evaluated the OOD detection performance of safe SSL methods using the F1 score, as shown in Supplementary Section S-2 (Table S-4). CaliMatch and OpenMatch outperformed other SSL methods in unseen-class detection across four datasets. In this section, we contrast the differences between CaliMatch and OpenMatch. To compare their unseen-class detection details, we used reliability diagrams \cite{niculescu2005predicting} to visualize the calibration performance, as shown in Figure \ref{image2}, with ECE metrics for each diagram. Our evaluation of unseen-label data detection calibration is based on considering OOD detection as the binary classification where the unseen class is recognized as a positive class.

The blue bar represents the average confidence scores provided by the models in each bin, while the red bar represents the accuracy of the OOD detection in each bin. If the blue bar is above the red one, the model overestimates the confidence in OOD detection. Therefore, OpenMatch's OOD detector generally overestimates the probabilities in OOD detection. This means that OpenMatch is more likely to accept unseen classes as seen classes or seen classes as unseen classes with unreasonably high confidence in OOD detection. On the other hand, CaliMatch provides better calibration performance across all four datasets, demonstrating that our approach to improving calibration performance worked as expected. Our proposed method's consistently lower ECE scores compared to those of OpenMatch across all datasets also suggest that the seen-class and OOD scores generated by our well-calibrated model are more effective in reliably selecting unlabeled seen-label samples during safe SSL.

To further investigate the importance of a well-calibrated OOD detector in safe SSL, we investigated how the performance of CaliMatch and OpenMatch varies as the decision threshold changes, as shown in Figure \ref{image3}. We increased the threshold $\tau_1$ from 0.5 to 0.8. The first graph in Figure \ref{image3} shows how the accuracy of the multiclass classification changes as a function of $\tau_{1}$. The accuracy of CaliMatch is higher than that of OpenMatch in all $\tau_{1}$ values, demonstrating that CaliMatch consistently outperforms OpenMatch with different OOD detection thresholds because of its superior calibration performance. The second graph provides the variations in the F1 score for unseen-class detection, selecting only observations with high-confidence values greater than $\tau_{1}$. Here, confidence refers to the higher value between the seen-class score and the OOD score. The last graph represents changes in the ratio of selected samples with high confidence among the seen-label data. As more uncertain predictions are removed, the F1 score increases, indicating that CaliMatch effectively avoids incorrect decisions and has a positive impact on safe SSL, thereby increasing accuracy as well, except when the value of $\tau_{1}$ changes from 0.7 to 0.8. This exception comes from the trade-off between the quality and quantity of selected seen-label samples. However, OpenMatch did not show any significant performance improvement in Figure \ref{image3}, even though it filtered out the uncertain decision results. Note that a more detailed analysis on $\tau_1$ and $\tau_2$ is also provided in Supplementary Section S-2.

\begin{table*}[]
\caption{Multiclass classification results of the ablation study on four datasets. The best results are in \textbf{bold}, and the second-best results are \underline{underlined}. (ACC: Accuracy, w/o: without)}
\centering
\resizebox{0.68\linewidth}{!}{
\begin{tabular}{c|cc|cc|cc|cc}
\hline
\multirow{2}{*}{Method}                                                                         & \multicolumn{2}{c|}{SVHN} & \multicolumn{2}{c|}{CIFAR-10} & \multicolumn{2}{c|}{CIFAR-100} & \multicolumn{2}{c}{TinyImageNet} \\ \cline{2-9} 
& ACC     & ECE        & ACC       & ECE          & ACC       & ECE          & ACC        & ECE            \\ \hline
\multirow{2}{*}{\begin{tabular}[c]{@{}c@{}}OpenMatch \\ (=w/o both calibration)\end{tabular}} &     95.67            &  0.026          &      86.19              & 0.115            &    65.65           &  0.256           & 44.02              &  0.413             \\
    & (0.14)            & (0.002)         & (0.74)              & (0.007)       & (1.02)             & (0.006)            & (0.63)              & (0.008)             \\ \hline
    
\multirow{2}{*}{\begin{tabular}[c]{@{}c@{}}without\\ multiclass calibration\end{tabular}}      & 96.08        & 0.022      & 86.47          & 0.108        & 66.16          & 0.223         & 43.68           & 0.341          \\ & (0.02)       & (0.002)    & (0.87)         & (0.008)      & (0.50)         & (0.006)       & (0.63)          & (0.020)        \\ \hline

\multirow{2}{*}{\begin{tabular}[c]{@{}c@{}}without \\ OOD calibration\end{tabular}}             & \underline{96.23}        & \underline{0.009}      & \underline{87.57 }         & \underline{0.038}        & \underline{66.96}          & \underline{0.033}         & \underline{44.65}           & \underline{0.250}          \\
& \underline{(0.09)}       & \underline{(0.004)}    & \underline{(0.22)}        & \underline{(0.010)}      & \underline{(0.30)}         &  \underline{(0.008)}       & \underline{(0.15)}          & \underline{(0.003)}        \\ \hline

\multirow{2}{*}{\textbf{CaliMatch}}                                          & \textbf{96.56}        & \textbf{0.006}      & \textbf{87.62}          & \textbf{0.029}        & \textbf{68.56}          & \textbf{0.025}         & \textbf{44.66}           & \textbf{0.233}          \\
& \textbf{(0.11)}       & \textbf{(0.002)}    & \textbf{(0.36)}        & \textbf{(0.003)}      & \textbf{(1.08)}         & \textbf{(0.007)}       & \textbf{(0.37)}          & \textbf{(0.006)}        \\ \hline
\end{tabular}
}
\label{table2}
\end{table*}

\begin{table*}[]
\caption{Unseen-label data detection results of the ablation study on four datasets. The best results are in \textbf{bold}, and the second-best results are \underline{underlined}. (w/o: without)}
\centering
\resizebox{0.68\linewidth}{!}{
\begin{tabular}{c|cc|cc|cc|cc}
\hline
\multirow{2}{*}{Method}                                                                         & \multicolumn{2}{c|}{SVHN} & \multicolumn{2}{c|}{CIFAR-10} & \multicolumn{2}{c|}{CIFAR-100} & \multicolumn{2}{c}{TinyImageNet} \\ \cline{2-9}                                                                  & F1     & ECE        & F1       & ECE          & F1       & ECE           & F1        & ECE            \\ \hline
\multirow{2}{*}{\begin{tabular}[c]{@{}c@{}}OpenMatch \\ (=w/o both calibration)\end{tabular}} 
 & 0.858 & 0.105 & 0.881 & 0.126 & \underline{0.696} & 0.338 & \underline{0.688} & 0.396 \\  
 & (0.009) & (0.004) & (0.013) & (0.006) & \underline{(0.002)} & (0.010) & \underline{(0.002)} & (0.006) \\ \hline

 \multirow{2}{*}{\begin{tabular}[c]{@{}c@{}}without \\ multiclass calibration\end{tabular}} 
 & \textbf{0.890} & \underline{0.037} & \textbf{0.909} & \underline{0.080} & 0.693 & \underline{0.149} & 0.683 & \textbf{0.189} \\  
 & \textbf{(0.004)} & \underline{(0.009)} & \textbf{(0.004)} & \underline{(0.010)} & (0.005) & \underline{(0.006)} & (0.004) & \textbf{(0.008)} \\ \hline

 \multirow{2}{*}{\begin{tabular}[c]{@{}c@{}}without \\ OOD calibration\end{tabular}} 
 & 0.865 & 0.085 & 0.858 & 0.086 & \textbf{0.709} & 0.320 & 0.686 & 0.366 \\  
 & (0.005) & (0.005) & (0.001) & (0.003) & \textbf{(0.001)} & (0.006) & (0.003) & (0.010) \\ \hline

 \multirow{2}{*}{\begin{tabular}[c]{@{}c@{}}\textbf{CaliMatch}\end{tabular}} 
 & \underline{0.889} & \textbf{0.036} & \underline{0.883} & \textbf{0.064} & 0.687 & \textbf{0.114} & \textbf{0.691} & \underline{0.238} \\  
 & \underline{(0.028)} & \textbf{(0.018)} & \underline{(0.003)} & \textbf{(0.005)} & (0.006) & \textbf{(0.008)} & \textbf{(0.001)} & \underline{(0.005)} \\ \hline
\end{tabular}
}
\label{table3}
\end{table*}

\subsection{Further Analyses}
Under $\kappa$ set to 60\% on four benchmark datasets, we conducted extensive ablation studies to demonstrate the effectiveness of calibration loss terms for the multiclass classifier and the OOD detector. We also compared OpenMatch with ablated CaliMatch approaches because OpenMatch is identical to CaliMatch when not calibrating all classifiers except for calculating the seen-class and OOD scores.

Table \ref{table2} presents that for multiclass classification, multiclass calibration had a more significant impact than OOD calibration across all datasets regarding both accuracy and ECE. In other words, our multiclass calibration is especially useful in mitigating overconfidence-based negative issues on SSL, such as low-quality pseudo-labels. Note that removing OOD calibration also resulted in slightly reduced classification and calibration performance for SVHN, CIFAR-10, and TinyImageNet while simultaneously causing a significant change in accuracy for CIFAR-100. It represents that the shared encoder $\phi_\theta$ of two models, $f_\theta \circ \phi_\theta$ and $g_\theta \circ \phi_\theta$, can also benefit from learning our adaptive smoothed labels of OOD detector regarding both accuracy and calibration improvements. Note that a Friedman test conducted across the four datasets also yielded statistically significant results ($p$-values $<$ 0.001), with the full CaliMatch configuration ranking first in both classification and calibration performance.

Table \ref{table3} shows that, for unseen-label detection, particularly in terms of calibration, OOD calibration had a greater impact than multiclass calibration across all datasets. It is noteworthy that not calibrating the multiclass classifier also resulted in significant increases in ECE across three datasets: CIFAR-10, CIFAR-100, and TinyImageNet. These results demonstrate the effectiveness of our calibration techniques in improving the calibration of unseen-label detection. Such calibration is a valuable asset in reliably selecting seen-label data, thereby positively impacting safe SSL, as illustrated in Figure \ref{image3}. In the case of TinyImageNet, only calibrating OOD detector resulted in the lowest calibration error for unseen-label detection compared to the original CaliMatch. This suggests that when the accuracies of both multiclass classifier and OOD detector are relatively poor, combining their predictions in CaliMatch's $s_i^u$ and $u_i^u$ may lead to more calibration errors in unseen-label detection compared to solely calibrating OOD detector, although CaliMatch already outperformed OpenMatch in terms of calibration. Additionally, there were minor fluctuations in the F1 score with $\tau_1$ set to 0.5 across all datasets, indicating that finding a proper threshold $\tau_1$ during calibration would be an additional task in safe SSL, as depicted in Figure \ref{image3}.

To highlight CaliMatch’s effectiveness on a more complex dataset, we compared it with OpenMatch and the baseline (supervised learning) on ImageNet with $\kappa$ set to 60\%. CaliMatch achieved a top-1 accuracy of 63.04$\pm$0.43\%, outperforming OpenMatch by 1.24\% and the baseline by 6.34\%. In terms of calibration in both classification and OOD detection, CaliMatch demonstrated ECE scores of 0.028 and 0.070, outperforming OpenMatch’s 0.041 and 0.121, respectively. These results highlight the critical role of calibration in improving SSL performance and CaliMatch's effectiveness for large-scale safe SSL scenarios.

To further examine the robustness and efficiency of CaliMatch, we performed two additional studies on CIFAR-10 with $\kappa$ set to 60\%. Detailed discussions are in Supplementary Section S-2 (Tables S-5 and S-6). We conducted a sensitivity analysis of $\lambda_{\text{O}}$ and $\lambda_{\text{OCal}}$, in which we did not observe a significant failure of our method in the range of hyperparameters considered. Moreover, we compared our adaptive calibration with other calibration methods in improving safe SSL. We found that while all of the calibration methods improved calibration when combined with OpenMatch, our method showed the highest improvement in calibration, ultimately leading to the best safe SSL performance.
\section{Conclusion}
This study proposed CaliMatch to address overconfidence-based negative issues in safe SSL scenarios. CaliMatch improved calibration of classification and OOD detection through adaptive label smoothing and scaling logit for multiclass and OvR classifiers. Our extensive experimental results with theoretical justification demonstrated that well-calibrated classifiers could enhance the safety and robustness of thresholding-based SSL methods by reducing negative training effects from incorrectly pseudo-labeled data whose confidence is high.

\section*{Acknowledgements}
The authors thank the ICCV reviewers for their careful evaluation and constructive suggestions. This research, conducted by Jinsoo Bae and Seoung Bum Kim, was supported by Brain Korea 21 FOUR, the Ministry of Science and ICT (MSIT) in Korea under the ITRC support program supervised by the Institute for Information Communication Technology Planning and Evaluation (IITP-2020-0-01749), and the National Research Foundation of Korea grant funded by the Korea government (RS-2022-00144190).
{
    \small
    \bibliographystyle{ieeenat_fullname}
    \bibliography{Reference}

\begin{thebibliography}{33}
\providecommand{\natexlab}[1]{#1}
\providecommand{\url}[1]{\texttt{#1}}
\expandafter\ifx\csname urlstyle\endcsname\relax
  \providecommand{\doi}[1]{doi: #1}\else
  \providecommand{\doi}{doi: \begingroup \urlstyle{rm}\Url}\fi

\bibitem[Bae et~al.(2022)Bae, Lee, and Kim]{bae2022safe}
Jinsoo Bae, Minjung Lee, and Seoung~Bum Kim.
\newblock Safe semi-supervised learning using a bayesian neural network.
\newblock \emph{Information Sciences}, 612:\penalty0 453--464, 2022.

\bibitem[Berthelot et~al.(2019)Berthelot, Carlini, Goodfellow, Papernot, Oliver, and Raffel]{berthelot2019mixmatch}
David Berthelot, Nicholas Carlini, Ian Goodfellow, Nicolas Papernot, Avital Oliver, and Colin~A Raffel.
\newblock Mixmatch: A holistic approach to semi-supervised learning.
\newblock \emph{Advances in Neural Information Processing Systems}, 32, 2019.

\bibitem[Berthelot et~al.(2020)Berthelot, Carlini, Cubuk, Kurakin, Sohn, Zhang, and Raffel]{2020remixmatch}
David Berthelot, Nicholas Carlini, Ekin~D. Cubuk, Alex Kurakin, Kihyuk Sohn, Han Zhang, and Colin Raffel.
\newblock Remixmatch: Semi-supervised learning with distribution alignment and augmentation anchoring.
\newblock In \emph{International Conference on Learning Representations}, 2020.

\bibitem[Chen et~al.(2020)Chen, Zhu, Li, and Gong]{chen2020semi}
Yanbei Chen, Xiatian Zhu, Wei Li, and Shaogang Gong.
\newblock Semi-supervised learning under class distribution mismatch.
\newblock In \emph{Proceedings of the AAAI Conference on Artificial Intelligence}, pages 3569--3576, 2020.

\bibitem[Chen et~al.(2023)Chen, Tan, Zhao, Chen, Song, Liang, and Lu]{chen2023boosting}
Yuhao Chen, Xin Tan, Borui Zhao, Zhaowei Chen, Renjie Song, Jiajun Liang, and Xuequan Lu.
\newblock Boosting semi-supervised learning by exploiting all unlabeled data.
\newblock In \emph{Proceedings of the IEEE/CVF Conference on Computer Vision and Pattern Recognition}, pages 7548--7557, 2023.

\bibitem[Deng et~al.(2009)Deng, Dong, Socher, Li, Li, and Fei-Fei]{deng2009imagenet}
Jia Deng, Wei Dong, Richard Socher, Li-Jia Li, Kai Li, and Li Fei-Fei.
\newblock Imagenet: A large-scale hierarchical image database.
\newblock In \emph{2009 IEEE conference on computer vision and pattern recognition}, pages 248--255. Ieee, 2009.

\bibitem[Guo et~al.(2017)Guo, Pleiss, Sun, and Weinberger]{guo2017calibration}
Chuan Guo, Geoff Pleiss, Yu Sun, and Kilian~Q Weinberger.
\newblock On calibration of modern neural networks.
\newblock In \emph{International Conference on Machine Learning}, pages 1321--1330. PMLR, 2017.

\bibitem[He et~al.(2016)He, Zhang, Ren, and Sun]{he2016deep}
Kaiming He, Xiangyu Zhang, Shaoqing Ren, and Jian Sun.
\newblock Deep residual learning for image recognition.
\newblock In \emph{Proceedings of the IEEE conference on computer vision and pattern recognition}, pages 770--778, 2016.

\bibitem[He et~al.(2022)He, Han, Lu, and Yin]{he2022safe}
Rundong He, Zhongyi Han, Xiankai Lu, and Yilong Yin.
\newblock Safe-student for safe deep semi-supervised learning with unseen-class unlabeled data.
\newblock In \emph{Proceedings of the IEEE/CVF Conference on Computer Vision and Pattern Recognition}, pages 14585--14594, 2022.

\bibitem[Kingma et~al.(2014)Kingma, Mohamed, Jimenez~Rezende, and Welling]{kingma2014semi}
Durk~P Kingma, Shakir Mohamed, Danilo Jimenez~Rezende, and Max Welling.
\newblock Semi-supervised learning with deep generative models.
\newblock \emph{Advances in Neural Information Processing Systems}, 27, 2014.

\bibitem[Krizhevsky et~al.(2009)Krizhevsky, Hinton, et~al.]{krizhevsky2009learning}
Alex Krizhevsky, Geoffrey Hinton, et~al.
\newblock Learning multiple layers of features from tiny images.
\newblock 2009.

\bibitem[Le and Yang(2015)]{le2015tiny}
Ya Le and Xuan Yang.
\newblock Tiny imagenet visual recognition challenge.
\newblock \emph{CS 231N}, 7\penalty0 (7):\penalty0 3, 2015.

\bibitem[Lee et~al.(2013)]{lee2013pseudo}
Dong-Hyun Lee et~al.
\newblock Pseudo-label: The simple and efficient semi-supervised learning method for deep neural networks.
\newblock In \emph{Workshop on Challenges in Representation Learning, ICML}, page 896. Atlanta, 2013.

\bibitem[Li et~al.(2023)Li, Qi, Shi, and Gao]{li2023iomatch}
Zekun Li, Lei Qi, Yinghuan Shi, and Yang Gao.
\newblock Iomatch: Simplifying open-set semi-supervised learning with joint inliers and outliers utilization.
\newblock In \emph{Proceedings of the IEEE/CVF International Conference on Computer Vision}, pages 15870--15879, 2023.

\bibitem[Liang et~al.(2018)Liang, Li, and Srikant]{liang2018enhancing}
Shiyu Liang, Yixuan Li, and R. Srikant.
\newblock Enhancing the reliability of out-of-distribution image detection in neural networks.
\newblock In \emph{International Conference on Learning Representations}, 2018.

\bibitem[Liu et~al.(2022)Liu, Ben~Ayed, Galdran, and Dolz]{liu2022devil}
Bingyuan Liu, Ismail Ben~Ayed, Adrian Galdran, and Jose Dolz.
\newblock The devil is in the margin: Margin-based label smoothing for network calibration.
\newblock In \emph{Proceedings of the IEEE/CVF Conference on Computer Vision and Pattern Recognition}, pages 80--88, 2022.

\bibitem[Minderer et~al.(2021)Minderer, Djolonga, Romijnders, Hubis, Zhai, Houlsby, Tran, and Lucic]{minderer2021revisiting}
Matthias Minderer, Josip Djolonga, Rob Romijnders, Frances Hubis, Xiaohua Zhai, Neil Houlsby, Dustin Tran, and Mario Lucic.
\newblock Revisiting the calibration of modern neural networks.
\newblock \emph{Advances in Neural Information Processing Systems}, 34:\penalty0 15682--15694, 2021.

\bibitem[M{\"u}ller et~al.(2019)M{\"u}ller, Kornblith, and Hinton]{muller2019does}
Rafael M{\"u}ller, Simon Kornblith, and Geoffrey~E Hinton.
\newblock When does label smoothing help?
\newblock \emph{Advances in Neural Information Processing Systems}, 32, 2019.

\bibitem[Naeini et~al.(2015)Naeini, Cooper, and Hauskrecht]{naeini2015obtaining}
Mahdi~Pakdaman Naeini, Gregory Cooper, and Milos Hauskrecht.
\newblock Obtaining well-calibrated probabilities using bayesian binning.
\newblock In \emph{Proceedings of the AAAI conference on artificial intelligence}, 2015.

\bibitem[Netzer et~al.(2011)Netzer, Wang, Coates, Bissacco, Wu, Ng, et~al.]{netzer2011reading}
Yuval Netzer, Tao Wang, Adam Coates, Alessandro Bissacco, Baolin Wu, Andrew~Y Ng, et~al.
\newblock Reading digits in natural images with unsupervised feature learning.
\newblock In \emph{NIPS workshop on deep learning and unsupervised feature learning}, page~7. Granada, Spain, 2011.

\bibitem[Niculescu-Mizil and Caruana(2005)]{niculescu2005predicting}
Alexandru Niculescu-Mizil and Rich Caruana.
\newblock Predicting good probabilities with supervised learning.
\newblock In \emph{International Conference on Learning Representations}, pages 625--632, 2005.

\bibitem[Noh et~al.(2023)Noh, Park, Lee, and Ham]{noh2023rankmixup}
Jongyoun Noh, Hyekang Park, Junghyup Lee, and Bumsub Ham.
\newblock Rankmixup: Ranking-based mixup training for network calibration.
\newblock In \emph{Proceedings of the IEEE/CVF International Conference on Computer Vision}, pages 1358--1368, 2023.

\bibitem[Oliver et~al.(2018)Oliver, Odena, Raffel, Cubuk, and Goodfellow]{oliver2018realistic}
Avital Oliver, Augustus Odena, Colin~A Raffel, Ekin~Dogus Cubuk, and Ian Goodfellow.
\newblock Realistic evaluation of deep semi-supervised learning algorithms.
\newblock \emph{Advances in Neural Information Processing Systems}, 31, 2018.

\bibitem[Saito et~al.(2021)Saito, Kim, and Saenko]{saito2021openmatch}
Kuniaki Saito, Donghyun Kim, and Kate Saenko.
\newblock Openmatch: Open-set semi-supervised learning with open-set consistency regularization.
\newblock \emph{Advances in Neural Information Processing Systems}, 34:\penalty0 25956--25967, 2021.

\bibitem[Sanchez~Aimar et~al.(2024)Sanchez~Aimar, Helgesen, Xu, Kuhlmann, and Felsberg]{sanchez2024flexible}
Emanuel Sanchez~Aimar, Nathaniel Helgesen, Yonghao Xu, Marco Kuhlmann, and Michael Felsberg.
\newblock Flexible distribution alignment: Towards long-tailed semi-supervised learning with proper calibration.
\newblock In \emph{European Conference on Computer Vision}, pages 307--327. Springer, 2024.

\bibitem[Sohn et~al.(2020)Sohn, Berthelot, Carlini, Zhang, Zhang, Raffel, Cubuk, Kurakin, and Li]{sohn2020fixmatch}
Kihyuk Sohn, David Berthelot, Nicholas Carlini, Zizhao Zhang, Han Zhang, Colin~A Raffel, Ekin~Dogus Cubuk, Alexey Kurakin, and Chun-Liang Li.
\newblock Fixmatch: Simplifying semi-supervised learning with consistency and confidence.
\newblock \emph{Advances in Neural Information Processing Systems}, 33:\penalty0 596--608, 2020.

\bibitem[Thulasidasan et~al.(2019)Thulasidasan, Chennupati, Bilmes, Bhattacharya, and Michalak]{thulasidasan2019mixup}
Sunil Thulasidasan, Gopinath Chennupati, Jeff~A Bilmes, Tanmoy Bhattacharya, and Sarah Michalak.
\newblock On mixup training: Improved calibration and predictive uncertainty for deep neural networks.
\newblock \emph{Advances in Neural Information Processing Systems}, 32, 2019.

\bibitem[Tomani et~al.(2021)Tomani, Gruber, Erdem, Cremers, and Buettner]{tomani2021post}
Christian Tomani, Sebastian Gruber, Muhammed~Ebrar Erdem, Daniel Cremers, and Florian Buettner.
\newblock Post-hoc uncertainty calibration for domain drift scenarios.
\newblock In \emph{Proceedings of the IEEE/CVF Conference on Computer Vision and Pattern Recognition}, pages 10124--10132, 2021.

\bibitem[Wang et~al.(2024)Wang, Xiang, Huang, Mao, Xiao, and Yamasaki]{wang2024scomatch}
Zerun Wang, Liuyu Xiang, Lang Huang, Jiafeng Mao, Ling Xiao, and Toshihiko Yamasaki.
\newblock Scomatch: Alleviating overtrusting in open-set semi-supervised learning.
\newblock In \emph{European Conference on Computer Vision}, pages 217--233. Springer, 2024.

\bibitem[Wei and Gan(2023)]{wei2023towards}
Tong Wei and Kai Gan.
\newblock Towards realistic long-tailed semi-supervised learning: Consistency is all you need.
\newblock In \emph{Proceedings of the IEEE/CVF Conference on Computer Vision and Pattern Recognition}, pages 3469--3478, 2023.

\bibitem[Yoon et~al.(2020)Yoon, Zhang, Jordon, and Van~der Schaar]{yoon2020vime}
Jinsung Yoon, Yao Zhang, James Jordon, and Mihaela Van~der Schaar.
\newblock Vime: Extending the success of self-and semi-supervised learning to tabular domain.
\newblock \emph{Advances in Neural Information Processing Systems}, 33:\penalty0 11033--11043, 2020.

\bibitem[Yu et~al.(2020)Yu, Ikami, Irie, and Aizawa]{yu2020multi}
Qing Yu, Daiki Ikami, Go Irie, and Kiyoharu Aizawa.
\newblock Multi-task curriculum framework for open-set semi-supervised learning.
\newblock In \emph{Computer Vision--ECCV 2020: 16th European Conference, Glasgow, UK, August 23--28, 2020, Proceedings, Part XII 16}, pages 438--454. Springer, 2020.

\bibitem[Zagoruyko and Komodakis(2016)]{BMVC2016_87}
Sergey Zagoruyko and Nikos Komodakis.
\newblock Wide residual networks.
\newblock \emph{Proceedings of the British Machine Vision Conference}, pages 87.1--87.12, 2016.

\end{thebibliography}


\begin{thebibliography}{14}
\providecommand{\natexlab}[1]{#1}
\providecommand{\url}[1]{\texttt{#1}}
\expandafter\ifx\csname urlstyle\endcsname\relax
  \providecommand{\doi}[1]{doi: #1}\else
  \providecommand{\doi}{doi: \begingroup \urlstyle{rm}\Url}\fi

\bibitem[Ajalloeian and Stich(2020)]{ajalloeian2020convergence}
Ahmad Ajalloeian and Sebastian~U Stich.
\newblock On the convergence of sgd with biased gradients.
\newblock \emph{arXiv preprint arXiv:2008.00051}, 2020.

\bibitem[Cubuk et~al.(2020)Cubuk, Zoph, Shlens, and Le]{cubuk2020randaugment}
Ekin~D Cubuk, Barret Zoph, Jonathon Shlens, and Quoc~V Le.
\newblock Randaugment: Practical automated data augmentation with a reduced search space.
\newblock In \emph{Proceedings of the IEEE/CVF conference on computer vision and pattern recognition workshops}, pages 702--703, 2020.

\bibitem[He et~al.(2016)He, Zhang, Ren, and Sun]{he2016deep}
Kaiming He, Xiangyu Zhang, Shaoqing Ren, and Jian Sun.
\newblock Deep residual learning for image recognition.
\newblock In \emph{Proceedings of the IEEE conference on computer vision and pattern recognition}, pages 770--778, 2016.

\bibitem[Huang et~al.(2017)Huang, Liu, Van Der~Maaten, and Weinberger]{huang2017densely}
Gao Huang, Zhuang Liu, Laurens Van Der~Maaten, and Kilian~Q Weinberger.
\newblock Densely connected convolutional networks.
\newblock In \emph{Proceedings of the IEEE Conference on Computer Vision and Pattern Recognition}, pages 4700--4708, 2017.

\bibitem[Li et~al.(2023)Li, Qi, Shi, and Gao]{li2023iomatch}
Zekun Li, Lei Qi, Yinghuan Shi, and Yang Gao.
\newblock Iomatch: Simplifying open-set semi-supervised learning with joint inliers and outliers utilization.
\newblock In \emph{Proceedings of the IEEE/CVF International Conference on Computer Vision}, pages 15870--15879, 2023.

\bibitem[Liu et~al.(2022)Liu, Ben~Ayed, Galdran, and Dolz]{liu2022devil}
Bingyuan Liu, Ismail Ben~Ayed, Adrian Galdran, and Jose Dolz.
\newblock The devil is in the margin: Margin-based label smoothing for network calibration.
\newblock In \emph{Proceedings of the IEEE/CVF Conference on Computer Vision and Pattern Recognition}, pages 80--88, 2022.

\bibitem[Noh et~al.(2023)Noh, Park, Lee, and Ham]{noh2023rankmixup}
Jongyoun Noh, Hyekang Park, Junghyup Lee, and Bumsub Ham.
\newblock Rankmixup: Ranking-based mixup training for network calibration.
\newblock In \emph{Proceedings of the IEEE/CVF International Conference on Computer Vision}, pages 1358--1368, 2023.

\bibitem[Oliver et~al.(2018)Oliver, Odena, Raffel, Cubuk, and Goodfellow]{oliver2018realistic}
Avital Oliver, Augustus Odena, Colin~A Raffel, Ekin~Dogus Cubuk, and Ian Goodfellow.
\newblock Realistic evaluation of deep semi-supervised learning algorithms.
\newblock \emph{Advances in Neural Information Processing Systems}, 31, 2018.

\bibitem[Saito et~al.(2021)Saito, Kim, and Saenko]{saito2021openmatch}
Kuniaki Saito, Donghyun Kim, and Kate Saenko.
\newblock Openmatch: Open-set semi-supervised learning with open-set consistency regularization.
\newblock \emph{Advances in Neural Information Processing Systems}, 34:\penalty0 25956--25967, 2021.

\bibitem[Sanchez~Aimar et~al.(2024)Sanchez~Aimar, Helgesen, Xu, Kuhlmann, and Felsberg]{sanchez2024flexible}
Emanuel Sanchez~Aimar, Nathaniel Helgesen, Yonghao Xu, Marco Kuhlmann, and Michael Felsberg.
\newblock Flexible distribution alignment: Towards long-tailed semi-supervised learning with proper calibration.
\newblock In \emph{European Conference on Computer Vision}, pages 307--327. Springer, 2024.

\bibitem[Sohn et~al.(2020)Sohn, Berthelot, Carlini, Zhang, Zhang, Raffel, Cubuk, Kurakin, and Li]{sohn2020fixmatch}
Kihyuk Sohn, David Berthelot, Nicholas Carlini, Zizhao Zhang, Han Zhang, Colin~A Raffel, Ekin~Dogus Cubuk, Alexey Kurakin, and Chun-Liang Li.
\newblock Fixmatch: Simplifying semi-supervised learning with consistency and confidence.
\newblock \emph{Advances in Neural Information Processing Systems}, 33:\penalty0 596--608, 2020.

\bibitem[Wang et~al.(2024)Wang, Xiang, Huang, Mao, Xiao, and Yamasaki]{wang2024scomatch}
Zerun Wang, Liuyu Xiang, Lang Huang, Jiafeng Mao, Ling Xiao, and Toshihiko Yamasaki.
\newblock Scomatch: Alleviating overtrusting in open-set semi-supervised learning.
\newblock In \emph{European Conference on Computer Vision}, pages 217--233. Springer, 2024.

\bibitem[Yu et~al.(2020)Yu, Ikami, Irie, and Aizawa]{yu2020multi}
Qing Yu, Daiki Ikami, Go Irie, and Kiyoharu Aizawa.
\newblock Multi-task curriculum framework for open-set semi-supervised learning.
\newblock In \emph{Computer Vision--ECCV 2020: 16th European Conference, Glasgow, UK, August 23--28, 2020, Proceedings, Part XII 16}, pages 438--454. Springer, 2020.

\bibitem[Zagoruyko and Komodakis(2016)]{BMVC2016_87}
Sergey Zagoruyko and Nikos Komodakis.
\newblock Wide residual networks.
\newblock \emph{Proceedings of the British Machine Vision Conference}, pages 87.1--87.12, 2016.

\end{thebibliography}
}
\end{document}


\clearpage
\setcounter{page}{1}
\maketitlesupplementary

\setcounter{table}{0}
\renewcommand{\thetable}{S-\arabic{table}}
\setcounter{figure}{0}
\renewcommand{\thefigure}{S-\arabic{figure}}
\setcounter{equation}{0}
\renewcommand{\theequation}{S-\arabic{equation}}
\setcounter{section}{0}
\renewcommand{\thesection}{S-\arabic{section}}

\section{Detailed Experiment Setups}
\label{Sup-A-1}
%

\begin{table*}[]
\caption{Label distribution and number of data on five benchmark datasets under safe SSL setup.}
\centering
\resizebox{\linewidth}{!}{
\begin{tabular}{c|ccc|ccc}
\hline
\multirow{2}{*}{Dataset} & \multicolumn{3}{c|}{Training labeled dataset}  & \multicolumn{3}{c}{Training unlabeled dataset} \\ \cline{2-7} 
 & \begin{tabular}[c]{@{}c@{}}Label\\ distribution\end{tabular} & \begin{tabular}[c]{@{}c@{}}Number of data\\ for each class\end{tabular} & \begin{tabular}[c]{@{}c@{}}Total\\ number of data\end{tabular} & \begin{tabular}[c]{@{}c@{}}Label\\ distribution\end{tabular} & \begin{tabular}[c]{@{}c@{}}Number of data\\ for each class\end{tabular} & \begin{tabular}[c]{@{}c@{}}Total\\ number of data\end{tabular} \\ \hline
SVHN                     & 2,3,4,5,6,7        & 50                                                                       & 300                                                            & 0-9                & 2,000                                                                   & 20,000                                                         \\
CIFAR-10                 & 2,3,4,5,6,7        & 400                                                                      & 2,400                                                          & 0-9                & 2,000                                                                   & 20,000                                                         \\
CIFAR-100                & 0-49               & 100                                                                      & 5,000                                                          & 0-99               & 200                                                                     & 20,000                                                         \\
TinyImageNet             & 0-99               & 100                                                                      & 10,000                                                         & 0-199              & 200                                                                     & 40,000                                                         \\ 
ImageNet             & 0-499               & 150                                                                      & 75,000                                                         & 0-999              & 500                                                                     & 500,000                                                         \\ 
\hline
\end{tabular}}
\label{table4}
\end{table*}

\begin{table*}[]
    \caption{Implementation details and hyperparameters on CIFAR-10, CIFAR-100, SVHN, and TinyImageNet.}
    \centering
\resizebox{0.7\linewidth}{!}{
   
        \begin{tabular}{lc}
        \hline
        \multicolumn{2}{c}{Shared}                                   \\ \hline
        Training iterations (epochs)                              & 500,000 (100) \\
        Iteration period of validation     & 5,000 \\
        Learning rate  & 0.003     \\
        Learning rate decay factor                         & 0.2     \\
        Learning rate decay at iteration                   & 400,000 \\ 
        Optimizer & Adam \\
        CNN backbone network & Wide ResNet 28-2 \cite{BMVC2016_87} \\
        Batch size for labeled and unlabeled data on SSL & 50 \\ \hline
        \multicolumn{2}{c}{Supervised learning}                      \\ \hline
        Batch size for labeled data & 100   \\ \hline
        \multicolumn{2}{c}{MTC}                            \\ \hline
        Parameters ($\alpha$,$\beta$) for the Beta distribution for mixup                                      & 0.75   \\
        Temperature parameter for sharpening in MixMatch                  & 0.5       \\
        Coefficient for mean squared error loss on unlabeled data                             & 75    \\ 
        OOD detector   & Single-layer perceptron    \\ 
        Pretraining iterations of OOD detector for stability & 50,000    \\ 
        Finetuning iterations of OOD detector and MixMatch                             & 450,000    \\ \hline
        \multicolumn{2}{c}{FixMatch}                                      \\ \hline
        Threshold for selecting unlabeled training data                                   & 0.95   \\
        Coefficient for cross-entropy loss on unlabeled data                   & 1     \\ \hline
        \multicolumn{2}{c}{OpenMatch}                             \\ \hline
        OOD detector                                      & Single-layer perceptron  \\
        Threshold for FixMatch   & 0.95    \\
        Threshold for selecting seen-label data   & 0.5    \\
        Coefficient for FixMatch's loss on unlabeled data                   & 1 \\
        Coefficient for entropy minimization of OOD detector                   & 0.1 \\
        Coefficient ($\lambda_{\text{S}}$) for soft open-set consistency regularization loss                  & 0.5       \\
        Warm-up training iterations for FixMatch                  & 25,000       \\ \hline
        
        \multicolumn{2}{c}{SafeStudent}                                      \\ \hline
        Temperature parameter for calculating energy discrepancy                               & \{1,1.5\}   \\
        Pretraining iterations for teacher model  & 100,000       \\
        Exponential moving average (EMA) factor & 0.996   \\ 
        Iteration period for the teacher model with EMA update & 50,000   \\ 
        Coefficient for confirmation bias elimination loss  & 1   \\ 
        Coefficient for unseen-class label distribution learning loss  & 0.01   \\ 
        \hline
        \multicolumn{2}{c}{IOMatch}                                     \\ \hline
        OOD detector & Single-layer perceptron  \\
        Open-set classifier  & Single-layer perceptron  \\
        Projection head & Three-layer perceptron  \\
        Threshold for FixMatch   & 0.95    \\ 
        Thresholds for selecting seen-label data & 0.5 \\ 
        Coefficients for FixMatch  & 1 \\ 
        Coefficients for multi-binary and open-set classifiers  & 1 \\ 
        \hline
        \multicolumn{2}{c}{SCOMatch}                        \\ \hline
        Size of OOD memory queue & max\{8$\times$number of classes,256\} \\ 
        Initial value for OOD detection and FixMatch  & 0.95 \\ 
        Positive and negative head classifier & Single-layer perceptron \\ 
        \hline
        \multicolumn{2}{c}{ADELLO}                        \\ \hline
        Exponential moving average decay & 0.999 \\ 
        Minimum of progressive alpha & 0.1 \\ 
        Value of progressive K & 2 \\ 
        Threshold for FixMatch & 0.95 \\ 
        \hline
        \multicolumn{2}{c}{\textbf{CaliMatch}}                        \\ \hline
        OOD detector                                      & Single-layer perceptron  \\
        Threshold ($\tau_2$) for FixMatch & 0.95    \\
        Threshold ($\tau_1$) for selecting seen-label data   & 0.5    \\
        Coefficient ($\lambda_{\text{O}}$) for classification loss of OOD detector                   & \{0.1,0.5,1\} \\
        Coefficient ($\lambda_{\text{OCal}}$) for calibration loss of OOD detector      & \{0.0005,0.001,0.1\}       \\
        Coefficient ($\lambda_{\text{S}}$) for soft open-set consistency regularization loss                  & 0.5       \\
        Epoch ($E_{\text{warm-up}}$) for warm-up stage in CaliMatch & 5 \\
        Number ($M$) of bins for adaptive label smoothing & 30 \\ \hline
        \end{tabular}}
    \label{table5}
\end{table*}

\paragraph{Datasets.} Table \ref{table4} summarizes the experimental settings of five datasets in this study under realistic SSL evaluation protocols \cite{oliver2018realistic} with label distribution mismatch scenarios. When we evaluated the SSL methods on multiclass classification in Tables 1 and 2 of the main paper, and Table \ref{table6}, we used only testing samples whose class labels align with the label distribution of the labeled dataset. On the other hand, when we evaluated the SSL methods on unseen-label data detection in Table 3 of the main paper and Table \ref{table7}, we used the original testing datasets, which contain all classes, including classes not seen from the label distribution of the labeled dataset. When we set the percentage of unseen-label data existence ($\kappa$) to 60\% in the unlabeled dataset, it indicates that 60\% of the training unlabeled samples come from unseen classes, while the rest of the samples come from seen classes.

\paragraph{Implementation for CIFAR-10, CIFAR-100, SVHN, and TinyImageNet.}
Table \ref{table5} summarizes the implementation details and hyperparameters used in our experiments on CIFAR-10, CIFAR-100, SVHN, and TinyImageNet. All hyperparameters listed in Table \ref{table5} were chosen based on achieving the highest average validation accuracy in multiclass classification or were adapted from previous studies.  In the validation stage, we used 10\% of the training dataset for each of the four datasets. We used standard normalization for scaling image data and several image augmentations, such as weak and strong augmentations, to implement consistency regularization on SSL methods. The weak augmentation used in this study includes horizontal-flip and random-crop, and the strong augmentation is grounded on RandAugment \cite{cubuk2020randaugment}. Additionally, when implementing other SSL methods for comparison with CaliMatch in our experimental setup, we referenced the official implementations provided by the authors: MTC \cite{yu2020multi}, FixMatch \cite{sohn2020fixmatch}, OpenMatch \cite{saito2021openmatch}, IOMatch \cite{li2023iomatch}, SCOMatch \cite{wang2024scomatch}, and ADELLO \cite{sanchez2024flexible}. To the best of our knowledge, there is no official code for the SafeStudent method; therefore, we directly implemented it on our experimental setup, and the source codes for SafeStudent and CaliMatch are distributed on our GitHub\textsuperscript{*}. All experiments were conducted using a single 12-GB GPU, such as an NVIDIA 3080 Ti.

\def\thefootnote{*}\footnotetext{https://github.com/bogus215/SafeSSL-Calibration.}

\paragraph{Implementation for ImageNet.}
For our ImageNet experiments, we trained CaliMatch, OpenMatch, and a supervised learning baseline using a distributed data-parallel approach on two NVIDIA A100 GPUs (80 GB) with 32 CPU cores. The backbone architecture is ResNet 50 \cite{he2016deep}, and training was conducted with a batch size of 160 labeled samples and 640 unlabeled samples per iteration. The models were trained for 200 epochs, following a learning rate schedule that included a linear warm-up phase over the first five epochs, gradually increasing to an initial learning rate of 0.4. Subsequently, the learning rate was decayed at epochs 60, 120, and 160, with a reduction factor of 0.1 at each step. Stochastic gradient descent (SGD) with Nesterov momentum (0.9) was used for optimization. In CaliMatch and OpenMatch, we applied two fixed thresholds ($\tau_1 = 0.5$ and $\tau_2 = 0.7$) used for OOD rejection and high quality of pseudo-labels. A weight decay coefficient of 0.0003 was used, incorporating L2 regularization on all model parameters. For data augmentation, we applied strong augmentation, including random color jitter, random grayscale, and random solarization, followed by an additional cutout operation. Weak augmentation was implemented using random horizontal flipping. During training, input images were randomly cropped and rescaled to 192×192, while during evaluation, images were center cropped and rescaled to 224×224, following standard ImageNet evaluation practices.

\section{Additional Results and Discussions}
\label{Sup-A-2}
\paragraph{Calibration in Classification.}
Table \ref{table6} presents the average ECE and standard deviation obtained from five runs on four datasets. All SSL methods exhibited enhanced calibration in comparison to supervised learning across the four testing datasets. This improvement stems from the enhanced accuracy achieved through SSL methods, which reduced the discrepancy between relatively high confidence and actual accuracy. However, most comparison SSL methods failed to surpass the calibration improvements achieved by CaliMatch, especially on SVHN, CIFAR-10, and CIFAR-100 datasets. This highlights CaliMatch's effectiveness in selecting high-accuracy data with high confidence, a valuable trait in thresholding-based SSL. In the case of MTC, it also showed improved calibration that was comparable to our method in some cases. This can be attributed to the smoothing-based calibration method, mixup, used in MixMatch for MTC, which enhanced the calibration of deep CNN. However, it did not consider the selection of unlabeled samples using its well-calibrated confidence despite its potential in SSL.

\begin{table*}[]
\caption{Evaluation of multiclass classification using the averaged ECE and standard deviation (in parentheses) on four image benchmark datasets under two different existence rates ($\kappa$\%) of unseen-label data. The best results are in \textbf{bold}, and the second-best results are \underline{underlined}.}
\label{table6}
\centering
\resizebox{0.9\textwidth}{!}{
\begin{tabular}{c|c|ccccccccc}
\hline
\multirow{2}{*}{Dataset}      & \multirow{2}{*}{$\kappa$} & \multicolumn{9}{c}{Method}                                                                                                                                                                                                                                                                                                                                                                                                                                                                                                                                                                                                                                                                                                                         \\ \cline{3-11} 
                              &                           & \multicolumn{1}{c|}{Supervised}                                                               & \multicolumn{1}{c|}{MTC}                                                              & \multicolumn{1}{c|}{FixMatch}                                                & \multicolumn{1}{c|}{OpenMatch}                                               & \multicolumn{1}{c|}{SafeStudent}                                             & \multicolumn{1}{c|}{IOMatch}                                                 & \multicolumn{1}{c|}{SCOMatch}                                                & \multicolumn{1}{c|}{ADELLO}                                                  & \textbf{CaliMatch}                                               \\ \hline
\multirow{2}{*}{\raisebox{-3mm}{SVHN}}         & 30\%                      & \multicolumn{1}{c|}{\multirow{2}{*}{\begin{tabular}[c]{@{}c@{}}\raisebox{-3mm}{0.203}\\ (0.008)\end{tabular}}} & \multicolumn{1}{c|}{\begin{tabular}[c]{@{}c@{}}0.054\\ (0.009)\end{tabular}}          & \multicolumn{1}{c|}{\begin{tabular}[c]{@{}c@{}}0.029\\ (0.002)\end{tabular}} & \multicolumn{1}{c|}{\begin{tabular}[c]{@{}c@{}}0.021\\ (0.002)\end{tabular}} & \multicolumn{1}{c|}{\begin{tabular}[c]{@{}c@{}}\underline{0.011}\\ \underline{(0.010)}\end{tabular}} & \multicolumn{1}{c|}{\begin{tabular}[c]{@{}c@{}}0.025\\ (0.005)\end{tabular}} & \multicolumn{1}{c|}{\begin{tabular}[c]{@{}c@{}}0.024\\ (0.006)\end{tabular}} & \multicolumn{1}{c|}{\begin{tabular}[c]{@{}c@{}}0.027\\ (0.008)\end{tabular}} & \textbf{\begin{tabular}[c]{@{}c@{}}0.003\\ (0.002)\end{tabular}} \\ \cline{2-2} \cline{4-11} 
                              & 60\%                      & \multicolumn{1}{c|}{}                                                                         & \multicolumn{1}{c|}{\begin{tabular}[c]{@{}c@{}}0.058\\ (0.008)\end{tabular}}          & \multicolumn{1}{c|}{\begin{tabular}[c]{@{}c@{}}0.036\\ (0.002)\end{tabular}} & \multicolumn{1}{c|}{\begin{tabular}[c]{@{}c@{}}0.026\\ (0.002)\end{tabular}} & \multicolumn{1}{c|}{\begin{tabular}[c]{@{}c@{}}\underline{0.016}\\ \underline{(0.010)}\end{tabular}} & \multicolumn{1}{c|}{\begin{tabular}[c]{@{}c@{}}0.032\\ (0.002)\end{tabular}} & \multicolumn{1}{c|}{\begin{tabular}[c]{@{}c@{}}0.033\\ (0.004)\end{tabular}} & \multicolumn{1}{c|}{\begin{tabular}[c]{@{}c@{}}0.038\\ (0.008)\end{tabular}} & \textbf{\begin{tabular}[c]{@{}c@{}}0.006\\ (0.002)\end{tabular}} \\ \hline
\multirow{2}{*}{\raisebox{-3mm}{CIFAR-10}}     & 30\%                      & \multicolumn{1}{c|}{\multirow{2}{*}{\begin{tabular}[c]{@{}c@{}}\raisebox{-3mm}{0.121}\\ (0.011)\end{tabular}}} & \multicolumn{1}{c|}{\begin{tabular}[c]{@{}c@{}}0.085\\ (0.057)\end{tabular}}          & \multicolumn{1}{c|}{\begin{tabular}[c]{@{}c@{}}0.097\\ (0.004)\end{tabular}} & \multicolumn{1}{c|}{\begin{tabular}[c]{@{}c@{}}0.096\\ (0.003)\end{tabular}} & \multicolumn{1}{c|}{\begin{tabular}[c]{@{}c@{}}0.107\\ (0.010)\end{tabular}} & \multicolumn{1}{c|}{\begin{tabular}[c]{@{}c@{}}0.083\\ (0.004)\end{tabular}} & \multicolumn{1}{c|}{\begin{tabular}[c]{@{}c@{}}\underline{0.076}\\ \underline{(0.006)}\end{tabular}} & \multicolumn{1}{c|}{\begin{tabular}[c]{@{}c@{}}0.091\\ (0.007)\end{tabular}} & \textbf{\begin{tabular}[c]{@{}c@{}}0.031\\ (0.004)\end{tabular}}          \\ \cline{2-2} \cline{4-11} 
                              & 60\%                      & \multicolumn{1}{c|}{}                                                                         & \multicolumn{1}{c|}{\begin{tabular}[c]{@{}c@{}}\underline{0.062}\\ \underline{(0.008)}\end{tabular}}          & \multicolumn{1}{c|}{\begin{tabular}[c]{@{}c@{}}0.116\\ (0.007)\end{tabular}} & \multicolumn{1}{c|}{\begin{tabular}[c]{@{}c@{}}0.115\\ (0.007)\end{tabular}} & \multicolumn{1}{c|}{\begin{tabular}[c]{@{}c@{}}0.123\\ (0.005)\end{tabular}} & \multicolumn{1}{c|}{\begin{tabular}[c]{@{}c@{}}0.107\\ (0.009)\end{tabular}} & \multicolumn{1}{c|}{\begin{tabular}[c]{@{}c@{}}0.109\\ (0.004)\end{tabular}} & \multicolumn{1}{c|}{\begin{tabular}[c]{@{}c@{}}0.106\\ (0.005)\end{tabular}} & \textbf{\begin{tabular}[c]{@{}c@{}}0.029\\ (0.003)\end{tabular}} \\ \hline
\multirow{2}{*}{\raisebox{-3mm}{CIFAR-100}}    & 30\%                      & \multicolumn{1}{c|}{\multirow{2}{*}{\begin{tabular}[c]{@{}c@{}}\raisebox{-3mm}{0.340}\\ (0.005)\end{tabular}}} & \multicolumn{1}{c|}{\begin{tabular}[c]{@{}c@{}}\underline{0.046}\\ \underline{(0.012)}\end{tabular}}          & \multicolumn{1}{c|}{\begin{tabular}[c]{@{}c@{}}0.234\\ (0.004)\end{tabular}} & \multicolumn{1}{c|}{\begin{tabular}[c]{@{}c@{}}0.226\\ (0.007)\end{tabular}} & \multicolumn{1}{c|}{\begin{tabular}[c]{@{}c@{}}0.161\\ (0.010)\end{tabular}} & \multicolumn{1}{c|}{\begin{tabular}[c]{@{}c@{}}0.216\\ (0.009)\end{tabular}} & \multicolumn{1}{c|}{\begin{tabular}[c]{@{}c@{}}0.264\\ (0.010)\end{tabular}} & \multicolumn{1}{c|}{\begin{tabular}[c]{@{}c@{}}0.233\\ (0.006)\end{tabular}} & \textbf{\begin{tabular}[c]{@{}c@{}}0.025\\ (0.008)\end{tabular}} \\ \cline{2-2} \cline{4-11} 
                              & 60\%                      & \multicolumn{1}{c|}{}                                                                         & \multicolumn{1}{c|}{\begin{tabular}[c]{@{}c@{}}\underline{0.083}\\ \underline{(0.036)}\end{tabular}}          & \multicolumn{1}{c|}{\begin{tabular}[c]{@{}c@{}}0.257\\ (0.012)\end{tabular}} & \multicolumn{1}{c|}{\begin{tabular}[c]{@{}c@{}}0.256\\ (0.006)\end{tabular}} & \multicolumn{1}{c|}{\begin{tabular}[c]{@{}c@{}}0.171\\ (0.010)\end{tabular}} & \multicolumn{1}{c|}{\begin{tabular}[c]{@{}c@{}}0.233\\ (0.007)\end{tabular}} & \multicolumn{1}{c|}{\begin{tabular}[c]{@{}c@{}}0.276\\ (0.010)\end{tabular}} & \multicolumn{1}{c|}{\begin{tabular}[c]{@{}c@{}}0.260\\ (0.007)\end{tabular}} & \textbf{\begin{tabular}[c]{@{}c@{}}0.025\\ (0.007)\end{tabular}} \\ \hline
\multirow{2}{*}{\raisebox{-3mm}{TinyImageNet}} & 30\%                      & \multicolumn{1}{c|}{\multirow{2}{*}{\begin{tabular}[c]{@{}c@{}}\raisebox{-3mm}{0.513}\\ (0.009)\end{tabular}}} & \multicolumn{1}{c|}{\textbf{\begin{tabular}[c]{@{}c@{}}0.120\\ (0.015)\end{tabular}}} & \multicolumn{1}{c|}{\begin{tabular}[c]{@{}c@{}}0.385\\ (0.007)\end{tabular}} & \multicolumn{1}{c|}{\begin{tabular}[c]{@{}c@{}}0.388\\ (0.006)\end{tabular}} & \multicolumn{1}{c|}{\begin{tabular}[c]{@{}c@{}}0.323\\ (0.010)\end{tabular}} & \multicolumn{1}{c|}{\begin{tabular}[c]{@{}c@{}}0.363\\ (0.008)\end{tabular}} & \multicolumn{1}{c|}{\begin{tabular}[c]{@{}c@{}}0.406\\ (0.008)\end{tabular}} & \multicolumn{1}{c|}{\begin{tabular}[c]{@{}c@{}}0.398\\ (0.006)\end{tabular}} & \begin{tabular}[c]{@{}c@{}}\underline{0.189}\\ \underline{(0.008)}\end{tabular} \\ \cline{2-2} \cline{4-11} 
                              & 60\%                      & \multicolumn{1}{c|}{}                                                                         & \multicolumn{1}{c|}{\textbf{\begin{tabular}[c]{@{}c@{}}0.133\\ (0.020)\end{tabular}}} & \multicolumn{1}{c|}{\begin{tabular}[c]{@{}c@{}}0.408\\ (0.006)\end{tabular}} & \multicolumn{1}{c|}{\begin{tabular}[c]{@{}c@{}}0.413\\ (0.008)\end{tabular}} & \multicolumn{1}{c|}{\begin{tabular}[c]{@{}c@{}}0.338\\ (0.010)\end{tabular}} & \multicolumn{1}{c|}{\begin{tabular}[c]{@{}c@{}}0.393\\ (0.007)\end{tabular}} & \multicolumn{1}{c|}{\begin{tabular}[c]{@{}c@{}}0.420\\ (0.009)\end{tabular}} & \multicolumn{1}{c|}{\begin{tabular}[c]{@{}c@{}}0.412\\ (0.009)\end{tabular}} & \begin{tabular}[c]{@{}c@{}}\underline{0.233}\\ \underline{(0.006)}\end{tabular} \\ \hline
\end{tabular}
}
\end{table*}

\begin{table*}[t]
  \caption{Evaluation of unseen-label data detection using the averaged F1 and standard deviation (in parentheses) on four datasets. The best results are in \textbf{bold}, and the second-best results are \underline{underlined}.}
  
\centering
\begin{tabular}{c|cccccccc}
\hline
\multirow{2}{*}{Dataset}      & \multicolumn{8}{c}{Method}                                                                            \\ \cline{2-9} 
                              & MTC     & FixMatch & OpenMatch        & SafeStudent & IOMatch & SCOMatch & ADELLO  & CaliMatch        \\ \hline
\multirow{2}{*}{SVHN}         & 0.701   & 0.200    & 0.858            & 0.676       & 0.118   & \underline{0.867}    & 0.188   & \textbf{0.889}   \\
                              & (0.068) & (0.023)  & (0.009)          & (0.003)     & (0.037) & \underline{(0.010)}  & (0.024) & \textbf{(0.028)} \\ \hline
\multirow{2}{*}{CIFAR-10}     & 0.700   & 0.140    & \underline{0.881}            & 0.695       & 0.156   & 0.503    & 0.175   & \textbf{0.883}   \\
                              & (0.029) & (0.021)  & \underline{(0.013)}          & (0.003)     & (0.063) & (0.008)  & (0.033) & \textbf{(0.003)}   \\ \hline
\multirow{2}{*}{CIFAR-100}    & 0.001   & 0.486    & \textbf{0.696}   & \underline{0.691}       & 0.419   & 0.385    & 0.442   & 0.687            \\
                              & (0.000) & (0.026)  & \textbf{(0.002)} & \underline{(0.002)}     & (0.013) & (0.016)  & (0.031) & (0.006)          \\ \hline
\multirow{2}{*}{TinyImageNet} & 0.001   & 0.581    & \underline{0.688}            & 0.682       & 0.627   & 0.372    & 0.570   & \textbf{0.691}   \\
                              & (0.000) & (0.008)  & \underline{(0.002)}          & (0.004)     & (0.039) & (0.008)  & (0.009) & \textbf{0.001}   \\ \hline
\end{tabular}
\label{table7}
\end{table*}

\paragraph{Unseen-label Data Detection.} Table \ref{table7} presents the average F1 and standard deviation of safe SSL methods to evaluate the unseen-label detection performance across five repeated runs with different random seeds. In the case of FixMatch and ADELLO, we defined their unseen-label score by subtracting confidence, which is the maximum probability value in multiclass classification, from one. Among the eight SSL methods, OpenMatch and CaliMatch showed the best or second-best performance across all datasets. MTC also showed satisfactory performance on SVHN and CIFAR-10, whose number of classes is relatively small when compared to CIFAR-100 and TinyImageNet. The F1 scores of MTC on CIFAR-100 and TinyImageNet indicate that MTC's OOD detector made incorrect decisions by assigning all testing samples to classes, which are in the label distribution of the labeled dataset. This suggests that training the OOD detector on datasets with a large number of similar classes can be unstable and may fail to distinguish between unknown classes and similarly known classes. When we implemented IOMatch using its official code in our experimental settings, we empirically observed that the projection head it utilized was sensitive to the performance of its OOD detector, thus negatively affecting unseen-label detection despite our efforts to find better hyperparameter settings for IOMatch.

\begin{figure*}[!hbt]
\centering
\subfloat[SVHN]{\includegraphics[width=0.7\textwidth]{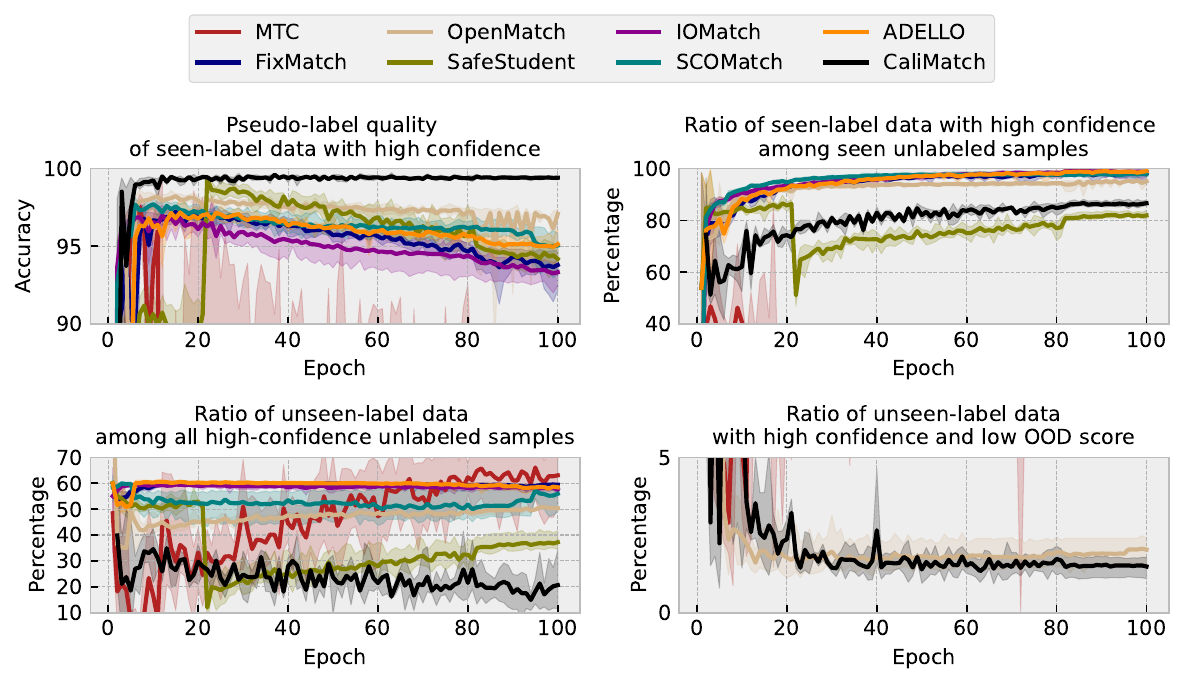}} \quad \quad 
\subfloat[CIFAR-10]{\includegraphics[width=0.7\textwidth]{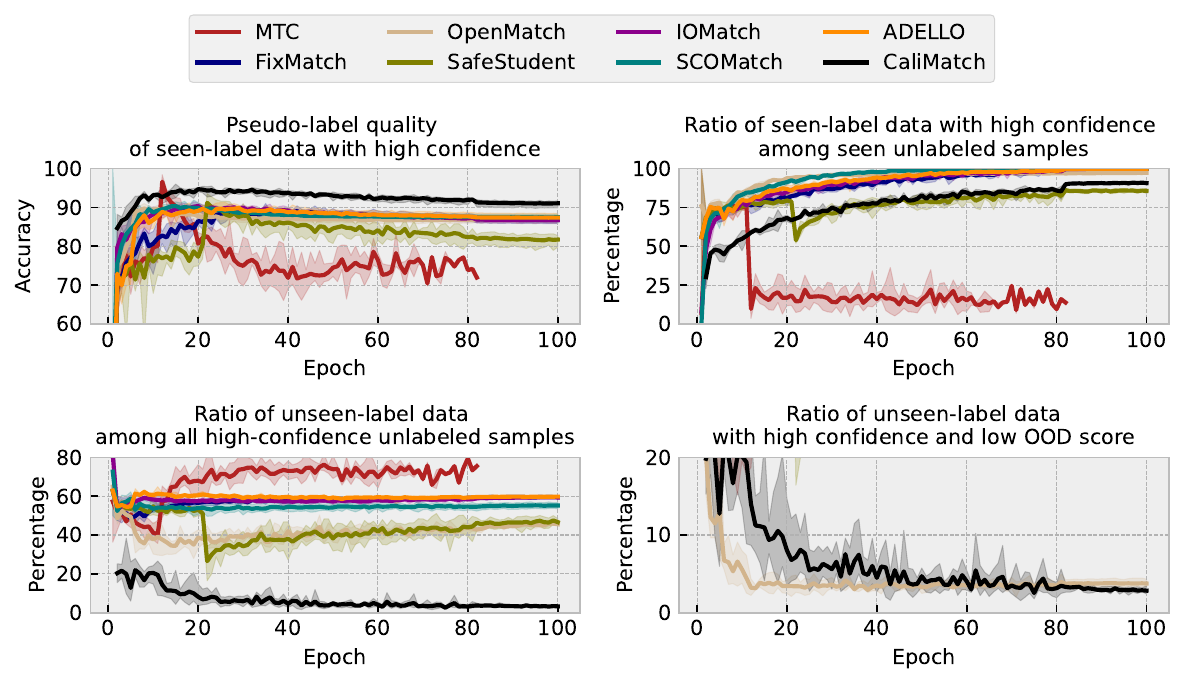}} \quad \quad 
\subfloat[TinyImageNet]{\includegraphics[width=0.7\textwidth]{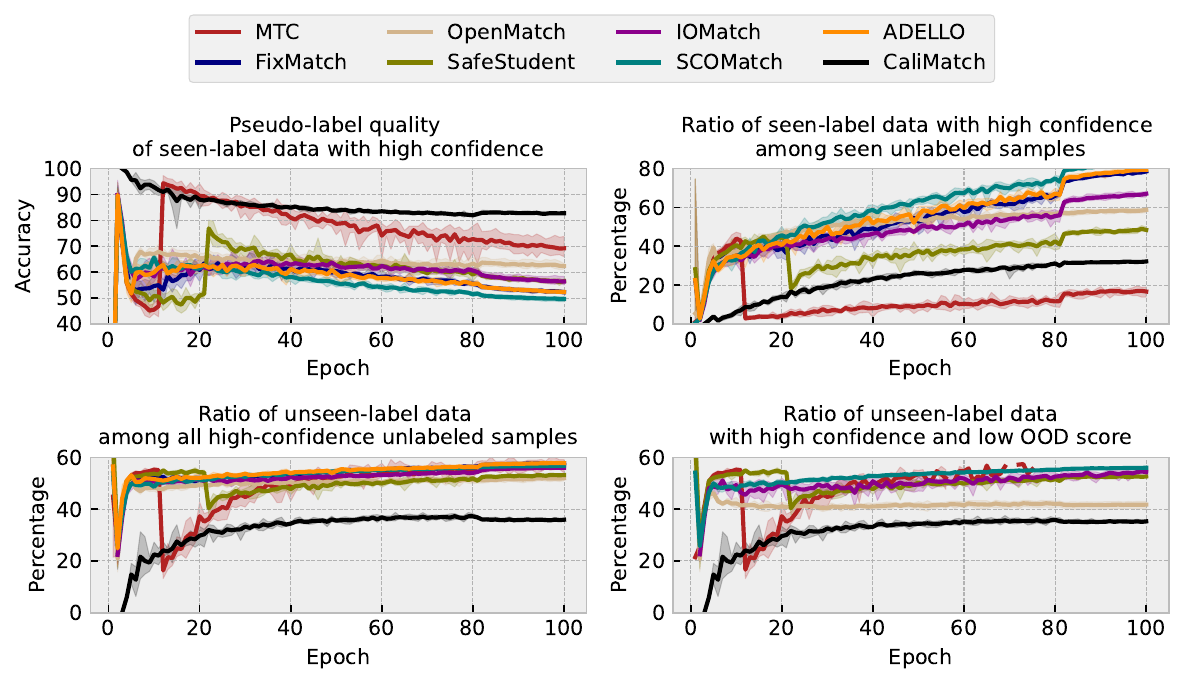}} \quad
\caption{Learning curves averaged over five runs on SVHN, CIFAR-10, and TinyImageNet for CaliMatch and other SSL methods. The shaded region indicates standard deviations calculated from five runs.} \label{image4}
\end{figure*}

\paragraph{Learning Curves.} To demonstrate our CaliMatch's calibration performance in terms of safe SSL, we also present the learning curves of all SSL methods on SVHN, CIFAR-10, and TinyImageNet in Figure \ref{image4}. In Figure \ref{image4}, we can have a similar discussion with Figure 1 of the main paper by demonstrating CaliMatch's superiority and robustness across all datasets. To highlight the SSL methods with the best or second-best performance in each plot of Figure \ref{image4}, we sometimes do not show the results of other SSL methods in detail if their differences compared to the best or second-best methods are significant. In the case of MTC on CIFAR-10, it exhibited instability in learning, failing to sustain long-term training. The training was interrupted around the 80th epoch by a gradient exploding in its OOD detector.

\paragraph{Sensitivity Analysis.} Table \ref{sensitivity-analysis} summarizes the results of sensitivity analysis on two hyperparameters ($\lambda_{\text{O}}$ and $\lambda_{\text{OCal}}$) in CaliMatch. We observed that the performance of safe SSL depends on the choice of $\lambda_{\text{O}}$ and $\lambda_{\text{OCal}}$, but no significant failures occurred within the range of hyperparameters considered in this analysis. This suggests that, while a naive choice of hyperparameters may not yield optimal performance, our method is robust enough to perform reliably in most cases. Furthermore, we confirmed that, across all hyperparameter combinations, CaliMatch consistently outperformed other safe SSL methods, such as OpenMatch and IOMatch, in the multiclass classification task.

\begin{table*}[!hbt]
\caption{Sensitivity analysis of $\lambda_{\text{O}}$ and $\lambda_{\text{OCal}}$ for CaliMatch on CIFAR-10 with $\kappa = 60\%$.}
\centering
\begin{tabular}{cc|cc|cc}
\hline
\multicolumn{2}{c|}{Coefficient}                                                                                     & \multicolumn{2}{c|}{Multiclass classification} & \multicolumn{2}{c}{Unseen-label detection} \\ \hline

$\lambda_{\text{O}}$ & $\lambda_{\text{OCal}}$ & Accuracy               & ECE                   & F1                   & ECE                 \\ \hline
\multirow{2}{*}{0.1}                                    & \multirow{2}{*}{0.1}                                       & 87.62                  & 0.029                 & 0.883                & 0.064               \\
                                                        &                                                            & (0.36)                 & (0.003)               & (0.003)              & (0.005)             \\ \hline
\multirow{2}{*}{0.1}                                    & \multirow{2}{*}{0.05}                                      & 87.94                  & 0.037                 & 0.881                & 0.066               \\
                                                        &                                                            & (0.57)                 & (0.010)               & (0.010)              & (0.013)             \\ \hline
\multirow{2}{*}{0.1}                                    & \multirow{2}{*}{0.01}                                      & 87.74                  & 0.041                 & 0.873                & 0.044               \\
                                                        &                                                            & (0.57)                 & (0.012)               & (0.006)              & (0.006)             \\ \hline
\multirow{2}{*}{0.5}                                    & \multirow{2}{*}{0.1}                                       & 87.51                  & 0.037                 & 0.875                & 0.074               \\
                                                        &                                                            & (0.33)                 & (0.013)               & (0.004)              & (0.014)             \\ \hline
\multirow{2}{*}{1}                                      & \multirow{2}{*}{0.1}                                       & 86.86                  & 0.039                 & 0.872                & 0.062               \\
                                                        &                                                            & (0.22)                 & (0.011)               & (0.006)              & (0.011)             \\ \hline
\end{tabular}
\label{sensitivity-analysis}
\end{table*}

\begin{table*}[!hbt]
\caption{Evaluation of multiclass classification and unseen-label detection using OpenMatch with calibration methods, as well as CaliMatch, on CIFAR-10 with $\kappa$ set to 60\%. The best results are in \textbf{bold}, and the second-best results are \underline{underlined}.}
\centering
\resizebox{0.9\textwidth}{!}{
\begin{tabular}{c|cc|cc|cc}
\hline
\multirow{2}{*}{Method}                                                                   & \multicolumn{2}{c|}{Calibration}           & \multicolumn{2}{c|}{Multiclass classification} & \multicolumn{2}{c}{Unseen-label detection} \\ \cline{2-7} 
                                                                                          & Multiclass classifier & OOD detector       & Accuracy               & ECE                   & F1                   & ECE                 \\ \hline
\multirow{2}{*}{OpenMatch}                                                                & \multirow{2}{*}{\ding{55}}    & \multirow{2}{*}{\ding{55}} & 86.19                  & 0.115                 & 0.881                & 0.126               \\
                                                                                          &                       &                    & (0.74)                 & (0.007)               & (0.013)              & (0.006)             \\ \hline
\multirow{4}{*}{\begin{tabular}[c]{@{}c@{}}OpenMatch with\\ label smoothing\end{tabular}} & \multirow{2}{*}{\ding{51}}    & \multirow{2}{*}{\ding{55}} & 86.84 & 0.082 & 0.854 & 0.121  \\
& & & (0.40) & (0.004)               & (0.008)              & (0.008)             \\
                                                                                          & \multirow{2}{*}{\ding{51}}    & \multirow{2}{*}{\ding{51}} & 83.49                  & 0.074                 & \textbf{0.887}                & 0.070               \\
                                                                                          &                       &                    & (0.86)                 & (0.017)               & \textbf{(0.015)}              & (0.014)             \\ \hline
\multirow{2}{*}{OpenMatch with mixup}                                                    & \multirow{2}{*}{\ding{51}}    & \multirow{2}{*}{\ding{51}} & 86.57                  & \underline{0.060}                 & 0.878                & \underline{0.065}               \\
                                                                                          &                       &                    & (0.36)                 & \underline{(0.025)}               & (0.024)              & \underline{(0.025)}             \\ \hline
\multirow{2}{*}{OpenMatch with MbLS}                                                    & \multirow{2}{*}{\ding{51}}    & \multirow{2}{*}{\ding{55}} & 86.60                  & 0.092                 & 0.866                & 0.108               \\
                                                                                          &                       &                    & (0.51)                 & (0.007)               & (0.011)              & (0.010)             \\ \hline
\multirow{2}{*}{OpenMatch with RankMixup}                                                    & \multirow{2}{*}{\ding{51}}    & \multirow{2}{*}{\ding{55}} & \underline{87.13}                  & 0.095                 & 0.829                & 0.147               \\
                                                                                          &                       &                    & \underline{(0.54)}                 & (0.008)               & (0.013)              & (0.014)             \\ \hline
\multirow{2}{*}{\textbf{CaliMatch}}                                                                & \multirow{2}{*}{\ding{51}}    & \multirow{2}{*}{\ding{51}} & \textbf{87.62}                  & \textbf{0.029}                 & \underline{0.883}                & \textbf{0.064}               \\
& & & \textbf{(0.36)} & \textbf{(0.003)} & \underline{(0.003)} & \textbf{(0.005)}             \\ \hline
\end{tabular}}
\label{openmatch with calibration}
\end{table*}

\paragraph{Safe SSL with Improved Calibration.}
Model calibration aims to align a model’s confidence scores with its actual accuracy, particularly when the model exhibits overconfidence or underconfidence. Effective calibration requires accurately assessing the discrepancy between the model’s predicted confidence and its true accuracy, allowing the appropriate level of calibration to be applied. However, existing smoothing-based calibration methods, such as label smoothing and mixup, typically rely on fine-grained grid search to determine the optimal smoothing level but still fail to generalize well across different data distributions. Liu et al. \cite{liu2022devil} introduced a constrained-optimization approach to label smoothing and proposed margin-based LS (MbLS) for improved calibration. MbLS relaxes the equality constraint used in standard label smoothing by enforcing a more flexible inequality constraint, leading to better calibration. However, MbLS applies a fixed smoothing intensity to all instances, failing to adapt to variations across different data regions. This limitation can result in suboptimal calibration because different data samples require different levels of calibration. Similarly, Noh et al. \cite{noh2023rankmixup} proposed RankMixup, which ranks mixup-augmented images and raw images based on their relative difficulty. Their method assumes that a model’s confidence in mixup-generated samples should be lower than in raw images, helping to improve calibration. However, mixup-based calibration heavily depends on a hyperparameter controlling the degree of interpolation, making it sensitive to tuning and potentially unreliable across diverse datasets.

In contrast, our calibration approach introduces a dynamic, adaptive mechanism that estimates the model’s current accuracy at each training epoch using a labeled validation set. This enables the model to determine the appropriate level of label smoothing without requiring manual hyperparameter tuning. We align the model's confidence distribution with the estimated accuracy through adaptively smoothed labels with $T_M$ and $T_O$. The two learnable parameters optimize themselves to stabilize the calibration process as the models learn the adaptively smoothed labels. This stable characteristic would be a valuable factor when it is applied to other frameworks. To support our claims, we performed additional experiments on CIFAR-10 with the 60\% mismatch ratio to compare our approach and other calibration methods. We applied various methods to improve calibration performance to OpenMatch. The results are summarized in Table \ref{openmatch with calibration}. We observed that all methods improved calibration performances compared to OpenMatch, but the improvement from our adaptive smoothing-based calibration was the best and eventually improved safe SSL. When the classic label smoothing was applied to both the multiclass classifier and OOD detector of OpenMatch, the OvR binary classifiers in the OOD detector exhibited instability because of gradient explosion, failing to sustain SSL training. This result highlights the importance of our learnable parameter $T_M$ in OOD calibration.

\begin{figure*}[]
\centering
\includegraphics[width=0.7\textwidth]{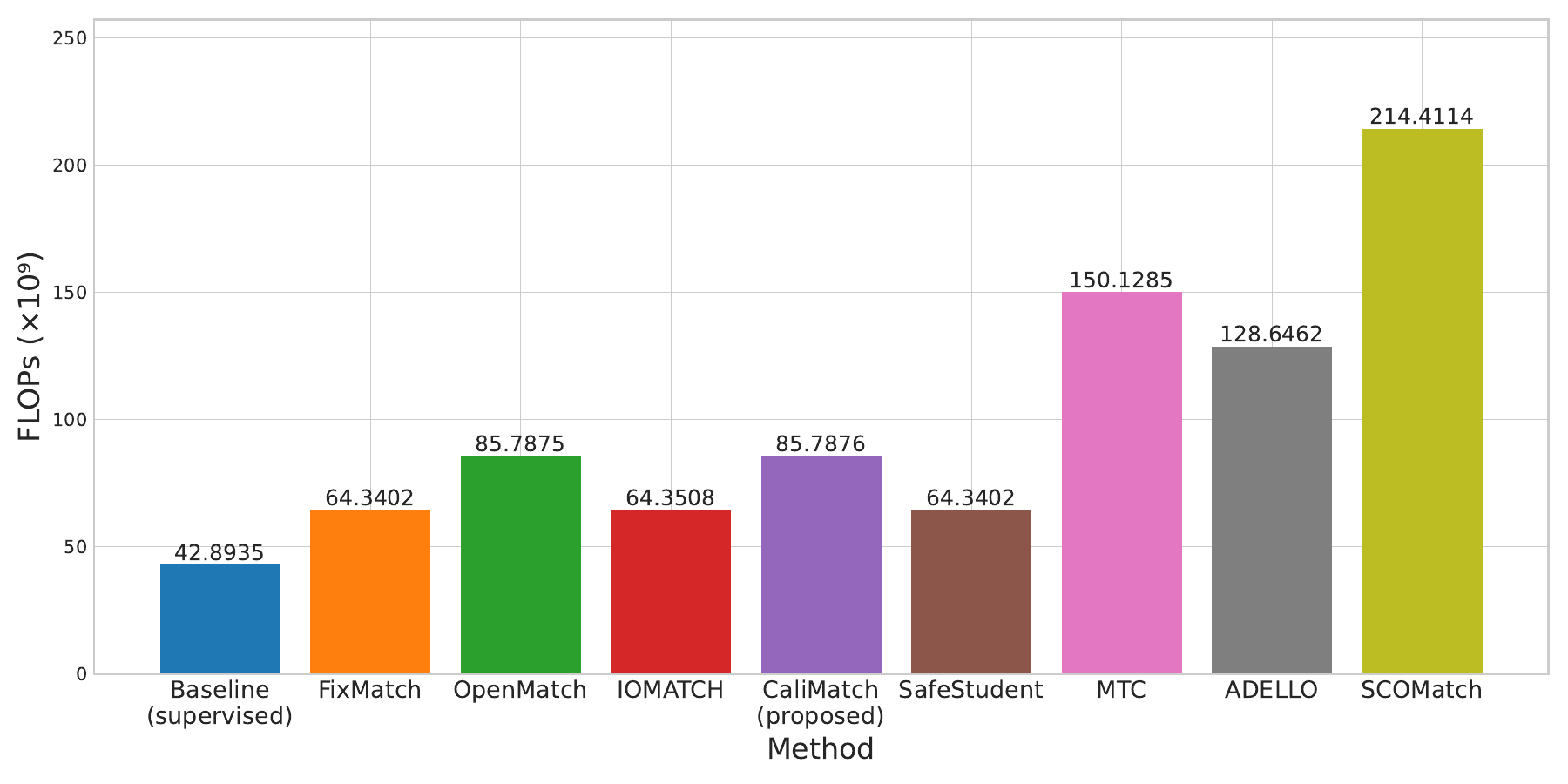}
\caption{Comparison of computational cost, measured in $10^9$ FLOPs, for SSL methods and a supervised baseline.}
\label{complexity}
\end{figure*}

\paragraph{Computational Complexity Analysis.}
To compare the computational complexities of the training methods, we calculate the number of floating point operations (FLOPs) required for one iteration of each approach. Note that the experimental setting for this analysis is on CIFAR-10 with $\kappa$ set to 60\%. As shown in Figure \ref{complexity}, the supervised learning baseline requires 42.89 giga FLOPs (GFLOPs, where 1 GFLOP = $10^9$ FLOPs). Among all the SSL methods, FixMatch, IOMatch, and SafeStudent have moderate computational costs of approximately 64.34 GFLOPs, while CaliMatch and OpenMatch exhibit slightly higher complexity at 85.79 GFLOPs. This is because the two methods use soft consistency regularization for OOD detectors based on two weakly augmented unlabeled images, demonstrating better OOD detection performance compared to the other SSL methods. Although CaliMatch incorporates additional techniques such as label smoothing and logit scaling, their computational impact is negligible, leading to no significant difference in FLOPs between CaliMatch and OpenMatch. In contrast, SCOMatch incurs the highest computational cost at 214.41 GFLOPs, significantly surpassing the other SSL methods. This is because SCOMatch learns from not only existing labeled and unlabeled data, but also new additional data labeled as OOD samples from their proposed OOD memory queue, resulting in increased computational overhead.

\paragraph{Thresholding in Safe SSL.} CaliMatch and OpenMatch have two threshold values $\tau_1$ and $\tau_2$ for safe SSL. Specifically, $\tau_1$ and $\tau_2$ are used to implement OOD rejection and FixMatch, respectively. On CIFAR-10 with $\kappa$ set to $60\%$, we investigated how the classification performance of CaliMatch and OpenMatch varies as the two threshold values change. As shown in Table \ref{taues}, although CaliMatch and OpenMatch exhibit some accuracy fluctuations on various $\tau_1$ and $\tau_2$ values, CaliMatch consistently achieves higher accuracy across all settings, particularly at higher thresholds.

\begin{table}[h]
\caption{Performance variations in CaliMatch and OpenMatch on CIFAR-10 with $\kappa$ set to 60\%. CaliMatch's results are in \textbf{bold}, and OpenMatch's results are in (parentheses).}
\centering
\resizebox{1\columnwidth}{!}{
\begin{tabular}{cc|ccc}
\hline
\multicolumn{2}{c|}{\multirow{2}{*}{Threshold}}     & \multicolumn{3}{c}{$\tau_2$}                                 \\ \cline{3-5} 
\multicolumn{2}{c|}{}                               & \multicolumn{1}{c|}{0.93} & \multicolumn{1}{c|}{0.95} & 0.97 \\ \hline
\multicolumn{1}{c|}{\multirow{4}{*}{$\tau_1$}} & 0.5 & \multicolumn{1}{c|}{\textbf{87.36} (86.13)}    & \multicolumn{1}{c|}{\textbf{87.62} (86.22)}    & \textbf{87.90} (86.12)    \\ \cline{2-5} 
\multicolumn{1}{c|}{}                         & 0.6 & \multicolumn{1}{c|}{\textbf{87.63} (85.80)}    & \multicolumn{1}{c|}{\textbf{87.62} (86.29)}    & \textbf{87.84} (86.19)   \\ \cline{2-5} 
\multicolumn{1}{c|}{}                         & 0.7 & \multicolumn{1}{c|}{\textbf{87.25} (85.56)}    & \multicolumn{1}{c|}{\textbf{88.13} (86.22)}    & \textbf{87.85} (86.54)   \\ \cline{2-5} 
\multicolumn{1}{c|}{}                         & 0.8 & \multicolumn{1}{c|}{\textbf{87.25} (86.27)}    & \multicolumn{1}{c|}{\textbf{87.72} (86.37)}    & \textbf{87.95} (85.92)   \\ \hline
\end{tabular}

}
\label{taues}
\end{table}

\paragraph{Backbone Network Variation.} To further evaluate the performance of CaliMatch with two additional CNN backbone networks (DenseNet‑121 \cite{huang2017densely} and ResNet‑50), we conducted experiments on CIFAR‑10 with $\kappa$ set to 60\%. All methods (CaliMatch, OpenMatch, and the supervised baseline) shared the identical hyperparameter configuration as those used in the experiments conducted for Table 1 of the main paper. As shown in Table \ref{backbone}, CaliMatch consistently outperforms both OpenMatch and the supervised approach in terms of accuracy, F1, and ECE across both backbone networks, demonstrating its robustness in the presence of unlabeled OOD samples during SSL. Notably, OpenMatch on DenseNet‑121 failed to surpass the supervised baseline, highlighting its inability to mitigate the detrimental effects of unlabeled OOD samples during SSL. These results underscore CaliMatch’s effectiveness and generality in handling safe SSL tasks across diverse backbone architectures.

\begin{table}[h]
\caption{Evaluation of multiclass classification and OOD detection of the baseline, OpenMatch, and CaliMatch with two popular CNNs on CIFAR-10 with $\kappa$ set to 60\%. The best results are highlighted in \textbf{bold}. (ACC: Accuracy)}
\centering
\resizebox{\columnwidth}{!}{
\begin{tabular}{c|c|cc|cc}
\hline
\multirow{2}{*}{Backbone}  & \multirow{2}{*}{Method} & \multicolumn{2}{c|}{Classification} & \multicolumn{2}{c}{OOD detection} \\
                           &                         & ACC           & ECE            & F1          & ECE           \\ \hline
\multirow{3}{*}{ResNet-50} & Baseline                & 56.41               & 0.409          & -                 & -             \\
                           & OpenMatch               & 58.72               & 0.307          & 0.740             & 0.131         \\
                           & \textbf{CaliMatch}               & \textbf{72.73}               & \textbf{0.123}          & \textbf{0.784}             & \textbf{0.120}         \\ \hline
\multirow{3}{*}{DenseNet-121}  & Baseline                & 75.20              & 0.226          & -                 & -             \\
                           & OpenMatch               & 71.60              & 0.224          & 0.782             & 0.189         \\
                           & \textbf{CaliMatch}               & \textbf{80.81}              & \textbf{0.070}          & \textbf{0.831}             & \textbf{0.096}         \\ \hline
\end{tabular}
}
\label{backbone}
\end{table}

\clearpage

\section{Theoretical Justification}
\label{Theory}
\subsection{Setups.}
Let $D_{u}^{a} = \{(x_{i}^{u}, y_{i}^{u}) \in \mathcal{X} \times \tilde{\mathcal{Y}}:i=1,\cdots,n_{u}\}$ be a set of unlabeled instances, where $\tilde{\mathcal{Y}}$ represents the set of all possible labels including the set of known labels $\mathcal{Y}$. We then consider $D_{u}^{s} = \{(x_{i}^{u},y^{u}_{i}) \in D_{u}^{a}:y_{i}^{u} \in \mathcal{Y}\}$, which 
contains only in-distribution (ID) samples. By successfully applying an OOD detection method on $D_{u}^{a}$, we may obtain $D_{u}^{s}$. Then we can calculate cross-entropy loss over $D_{u}^{s}$ with strong augmentations $\mathcal{T}_s$ for classification tasks:
\begin{equation}\label{eqn:ideal-loss}
    \mathcal{L}_{\text{CE}}(D^{s}_u;\mathcal{T}_s)=-\sum\limits_{i=1}^{|D^{s}_u|}\sum\limits_{k=1}^K y_{ik}^{u} \log p_{k}( \mathcal{T}_{s}(x_{i}^{u})).
\end{equation}
Training a classifier by minimizing $\mathcal{L}_{\text{CE}}(D_{u}^{s};\mathcal{T}_{s})$ as well as the cross-entropy loss on the labeled training dataset will provide further improvement compared to the case when we only consider cross-entropy loss on the labeled training dataset. However, it is challenging to identify the ``ideal'' dataset $D_{u}^{s}$ under safe SSL scenarios.

The safe SSL methods try to address this problem by having an OOD detector that identifies and discards instances from unseen classes within the unlabeled dataset $D_u$. They approximate the true labels $y_i^u$ for the remaining unlabeled instances in $D_u$ through pseudo-labeling techniques, which assign pseudo labels to unlabeled instances based on the model's predictions. In practice, these methods construct a surrogate loss function, $\mathcal{L}_{\text{Fix}}(B_u^t;\mathcal{T}_w,\mathcal{T}_s)$ (as seen in CaliMatch), which is designed to approximate the ideal loss $\mathcal{L}_{\text{CE}}(D_u^s;\mathcal{T}_s)$. One key consideration for the threshold-based safe SSL methods is that some unlabeled ID samples, despite having correct pseudo-labels, may be rejected from the surrogate loss computation because of low confidence. Here, we provide a theoretical analysis demonstrating how improving the calibration for both classification and OOD detection in safe SSL facilitates the alignment of gradients between the surrogate loss (provided by safe SSL) and the ideal loss (Equation \eqref{eqn:ideal-loss}), computed on the subset of samples that satisfy the thresholds. This alignment is crucial because, under stochastic gradient descent, surrogate gradients that closely approximate the ideal gradients can lead to comparable optimization outcomes \cite{ajalloeian2020convergence}. In other words, approximating the gradient of the ideal loss is sufficient to achieve a training effect similar to that obtained if the ideal loss were used directly.

\subsection{On the Importance of Calibration in Classification and OOD Detection.}
\newtheorem{theorem}{Theorem}
\newtheorem{corollary}{Corollary}
\newtheorem{lemma}{Lemma}

The goal of safe SSL is to ensure the quality of pseudo-labeling and the accuracy of OOD rejection, particularly for samples that meet the confidence threshold and OOD rejection threshold. The following lemma establishes that improving model calibration reduces the probability of incorrect pseudo-labeling or the inclusion of OOD samples in the training set $B_u^t$.

\begin{lemma}\label{lemma:high_confidence}
Assume the model is well-calibrated in the sense that for any confidence level \(s \in [0,1]\), the empirical accuracy of samples with predicted confidence in an interval \([s-\delta, s+\delta]\) is approximately \(s\) (with an error at most \(\eta\) for sufficiently small \(\delta>0\)). Here, we define $\varepsilon$ as follows:
\[
\varepsilon = P\Bigg( (x_i^u \in B_u^t) \wedge \Big( (y_i^u \notin \mathcal{Y}) \vee (y_i^u \neq \hat{y}_i^u) \Big) \Bigg),
\]
which represents the probability that a sample $x_i^u$ is either OOD or incorrectly pseudo-labeled. If the thresholds satisfy $\min\{\tau_1, \tau_2\} \ge 1-\eta$, then with probability at least \(1-\varepsilon\), any \(x_i^u\) from $B_u^t$ belongs to an ID class and is assigned the correct pseudo label:
\[
P\Big( (x_i^u\in B_u^t) \Rightarrow (y_i^u \in \mathcal{Y}) \wedge (y_i^u = \hat{y}_i^u) \Big) \geq 1 - \varepsilon.
\]
Moreover, as model calibration improves (i.e., as calibration error $\eta$ decreases), the probability of incorrect selection \(\varepsilon\) decreases.
\end{lemma}

This lemma implies that with better calibration, the number of incorrectly pseudo-labeled or OOD samples in the training set is minimized, ensuring that the threshold-based selection process primarily retains correctly pseudo-labeled ID samples.

\begin{proof}
A well-calibrated model satisfies that for any interval \(A\subset [0,1]\) of predicted confidence scores, the empirical accuracy on \(A\) is approximately equal to the mean confidence over \(A\). Formally, if $A=\{x\mid s(x)\in[s-\delta,s+\delta]\}$,
then, $\Big|P\big(y=\hat{y}\mid x\in A\big) - \mathbb{E}[s(x)\mid x \in A] \Big| \le \eta$. For samples $x_i^u$ in $B_u^t$, the empirical accuracy of pseudo-label is at least \(\tau_2-\eta\). In a similar way, by the calibration assumption, the empirical OOD detection accuracy for selected samples is at least $\tau_1 - \eta$. Hence, the probability that a sample $x_i^u$ is either misclassified or OOD is bounded by
\[
\varepsilon \le 1 - \min\{\tau_1, \tau_2\} + \eta.
\]
Thus, when the thresholds $\tau_1$ and $\tau_2$ are sufficiently high and the model is well-calibrated (i.e., $\eta$ is small), the probability of incorrect selection \(\varepsilon\) is minimized.
\end{proof}

Next, we present a theorem that establishes how improved calibration facilitates the alignment of gradients between the surrogate and ideal loss functions. This alignment is crucial in ensuring that safe SSL optimization behaves similarly to the supervised learning.

\begin{theorem}\label{thm:alignment}
Let \(\mathcal{L}_{\text{Fix}}(B_u^t; \mathcal{T}_w, \mathcal{T}_s)\) be the FixMatch-based loss on \(B_u^t\), and let \(\mathcal{L}_{\text{CE}}(B_u^s; \mathcal{T}_s)\) be the ideal cross-entropy loss computed on all ID unlabeled data \(B_u^s\) sampled from \(D_u^s\) with true labels. Define \( B_u^t \cap B_u^s \) as the subset of ID samples in \( B_u^t \), i.e., the correctly classified ID samples that meet the thresholds. Then, under the assumption that \(\varepsilon\) is sufficiently small, the gradient difference satisfies:
\[
\bigl\| \nabla_{\theta} \mathcal{L}_{\text{Fix}}(B_u^t; \mathcal{T}_w, \mathcal{T}_s) - \nabla_{\theta} \mathcal{L}_{\text{CE}}(B_u^t \cap B_u^s; \mathcal{T}_s) \bigr\| \leq C \varepsilon |B_u^t|,
\]
where \(C\) is a positive constant.
\end{theorem}

This theorem asserts that as the calibration error decreases, the gradients of the surrogate loss function become increasingly similar to those of the ideal loss, leading to more stable and reliable optimization in safe SSL settings.

\begin{proof}
Let \( f_\theta(\mathcal{T}_s(x)) \) produce logits \( z_i = [z_{i,1}, \dots, z_{i,K}] \) for \( K \) classes, and denote \( p_i = \mathrm{softmax}(z_i) \) as the predicted probabilities. The ideal cross-entropy loss for an ID sample \((x_i^u, y_i^u) \in B_u^s\) is given by:
\[
\ell_{\text{CE}}(z_i, y_i^u) = -\log p_{i,y_i^u}.
\]
For a sample \( x_i^u \in B_u^t \) with pseudo-label \( \hat{y}_i^u \), the surrogate loss is:
\[
\ell_{\text{CE}}(z_i, \hat{y}_i^u) = -\log p_{i,\hat{y}_i^u}.
\]
\paragraph{Step 1: Gradient Expression at Sample Level}
The gradient of the loss with respect to the logits is given by:
\[
\frac{\partial \ell_{\text{CE}}(z_i, y)}{\partial z_{i,k}} = p_{i,k} - \mathbb{I}(y = k).
\]
For correctly pseudo-labeled samples (i.e., \( \hat{y}_i^u = y_i^u \)), we have:
\[
\frac{\partial \ell_{\text{CE}}(z_i, \hat{y}_i^u)}{\partial z_{i,k}} = \frac{\partial \ell_{\text{CE}}(z_i, y_i^u)}{\partial z_{i,k}},
\]
meaning their gradient contributions to the surrogate and ideal losses are identical. For incorrectly pseudo-labeled samples (\( \hat{y}_i^u \neq y_i^u \)), the gradient error per sample is:

\[
\Delta g_i = \frac{\partial \ell_{\text{CE}}(z_i, \hat{y}_i^u)}{\partial z_{i,k}} - \frac{\partial \ell_{\text{CE}}(z_i, y_i^u)}{\partial z_{i,k}}.
\]
Since the difference between any two softmax gradients is bounded, there exists a constant \( B \) such that:
\[
\|\Delta g_i\| \leq B.
\]

\paragraph{Step 2: Batch-level Gradient Analysis}
Now, we analyze the batch-level gradient over \( B_u^t \) and compare it to the ideal batch gradient over \( B_u^t \cap B_u^s \). The surrogate gradient over a batch \( B_u^t \) is:
\[
\nabla_{\theta} \mathcal{L}_{\text{Fix}}(B_u^t;\mathcal{T}_w,\mathcal{T}_s)=\sum_{x_i^u \in B_u^t} \nabla_{\theta} \ell_{\text{CE}}(z_i, \hat{y}_i^u).
\]
The ideal gradient over a batch \( B_u^t \cap B_u^s \) is:
\[
\nabla_{\theta} \mathcal{L}_{\text{CE}}(B_u^t \cap B_u^s; \mathcal{T}_s)=\sum_{x_i^u \in B_u^t \cap B_u^s} \nabla_{\theta} \ell_{\text{CE}}(z_i, y_i^u).
\]
By Lemma~\ref{lemma:high_confidence}, with probability at least \( 1 - \varepsilon \), a sample in \( B_u^t \) is correctly pseudo-labeled, meaning that for these samples, the gradient difference is zero. However, for at most \( \varepsilon \)-fraction of the samples, the gradient error per sample is at most \( B \). Thus, the total batch-wise gradient difference satisfies:
\[
\bigl\| \nabla_{\theta} \mathcal{L}_{\text{Fix}}(B_u^t) - \nabla_{\theta} \mathcal{L}_{\text{CE}}(B_u^t \cap B_u^s) \bigr\| \leq B \varepsilon |B_u^t|.
\]
\paragraph{Step 3: Extending to Model Parameters}
Using the chain rule, since the Jacobian of the model parameters \( \frac{\partial z}{\partial \theta} \) has a bounded norm \( L \), the parameter-wise gradient difference satisfies:
\[
\bigl\| \nabla_{\theta} \mathcal{L}_{\text{Fix}}(B_u^t) - \nabla_{\theta} \mathcal{L}_{\text{CE}}(B_u^t \cap B_u^s) \bigr\| \leq L B \varepsilon |B_u^t|.
\]
Setting \( C = L B \) completes the proof.
\end{proof}

\clearpage
{
    \renewcommand{\refname}{S-4. References}
    \small
    \bibliographystyle{ieeenat_fullname}
    \bibliography{supp_bib}
}